%% file: root.tex
\documentclass[letterpaper, 10 pt, journal]{Packages/IEEEtran}
\IEEEoverridecommandlockouts                              % This command is only needed if 
\usepackage{times}
\usepackage{epsfig}
\usepackage{graphicx}
\usepackage{amsmath}
\usepackage{amssymb}
\usepackage{subfigure} 
\usepackage[ruled]{algorithm}
\usepackage{stackengine}
\usepackage{scalerel}
\usepackage{textcomp}
\usepackage{stfloats}
\usepackage{lettrine}
\usepackage{url}
\usepackage{soul}
\usepackage[ruled]{algorithm}
\usepackage{algpseudocode}
\usepackage{ctable}
\usepackage{cite}
\usepackage[export]{adjustbox}
\usepackage{import}
\usepackage{pgf}
\usepackage[bookmarks=true]{hyperref}

\makeatletter
\let\NAT@parse\undefined
\makeatother
\usepackage[numbers]{natbib}

\usepackage{Packages/sparo_acronyms}
\usepackage{Packages/sparo_math}
\usepackage{Packages/sparo_SIunits}
\usepackage{Packages/sparo_misc}
\usepackage{multirow}

\usepackage{soul,color}
\usepackage{verbatim} % for multiline comment

\usepackage{xcolor}
\newcommand{\bl}[1]{{\textcolor{blue}{#1}}}

\definecolor{gl}{HTML}{008000}
\newcommand{\gl}[1]{{\textcolor{gl}{#1}}}

\DeclareMathOperator*{\argmin}{argmin}

\usepackage[symbol]{footmisc}

\title{Narrowing your FOV with SOLiD: Spatially Organized and Lightweight Global Descriptor for FOV-constrained LiDAR Place Recognition}

\author{Hogyun Kim${}^{1}$, Jiwon Choi${}^{1}$, Taehu Sim${}^{2}$, Giseop Kim${}^{3}$, and Younggun Cho${}^{1*}$
\thanks{Manuscript received: April, 14, 2024; revised July, 2, 2024; accepted July, 21, 2024. This letter was recommended for publication by Associate Editor J. Civera upon evaluation of the reviewers' comments. This work was supported by the National Research Foundation of Korea(NRF) grant (No. RS-2023-00302589 and No.2022R1A4A3029480) and Institute of Information \& communications Technology Planning \& Evaluation grant (No.2022-0-00448) funded by the Korea government(MSIT). } %Use only for final RAL version
\thanks{${}^{1}$ H. Kim, ${}^{1}$ J. Choi, and ${}^{1*}$ Y. Cho are with the Dept. Electr. and Comput. Eng., Inha University, South Korea \texttt{(e-mail: [hg.kim, jiwon2]@inha.edu, yg.cho@inha.ac.kr)} ${}^{2}$ T. Sim is with the Dept. Electr. Eng., Inha University, South Korea \texttt{(e-mail: xogn2359@gmail.com)} ${}^{3}$ G. Kim is with Vision Group of NAVER LABS, Seongnam, Gyeonggi-do, 13561, Republic of Korea \texttt{(e-mail: giseop.kim@naverlabs.com)}}
\thanks{Digital Object Identifier (DOI): see top of this page.}
}

\begin{document}

% make the title area
\maketitle

\input{PaperWriting/abstract.tex}
\input{PaperWriting/introduction2.tex}
\input{PaperWriting/relatedwork.tex} 
\input{PaperWriting/method.tex}
\input{PaperWriting/experiments2.tex}
% \input{PaperWriting/limitation.tex}
\input{PaperWriting/conclusion.tex}

\scriptsize
\bibliographystyle{Packages/IEEEtranN} % not IEEEtran, but IEEEtranN for using cite author
\bibliography{Packages/string-short, Packages/references}

% that's all folks
\end{document}

%% file: PaperWriting/abstract.tex
\begin{abstract}
% 첫 문장 고치기
We often encounter limited FOV situations due to various factors such as sensor fusion or sensor mount in real-world robot navigation.
However, the limited FOV interrupts the generation of descriptions and impacts place recognition adversely.
Therefore, we suffer from correcting accumulated drift errors in a consistent map using LiDAR-based place recognition with limited FOV.
Thus, in this paper, we propose a robust LiDAR-based place recognition method for handling narrow FOV scenarios.
The proposed method establishes spatial organization based on the range-elevation bin and azimuth-elevation bin to represent places. 
In addition, we achieve a robust place description through reweighting based on vertical direction information.
Based on these representations, our method enables addressing rotational changes and determining the initial heading. Additionally, we designed a lightweight and fast approach for the robot's onboard autonomy. For rigorous validation, the proposed method was tested across various LiDAR place recognition scenarios (i.e., single-session, multi-session, and multi-robot scenarios).
To the best of our knowledge, we report the first method to cope with the restricted FOV.
Our place description and SLAM codes will be released.
Also, the supplementary materials of our descriptor are available at \texttt{\url{https://sites.google.com/view/lidar-solid}}.
\end{abstract}


%% file: PaperWriting/introduction2.tex
\section{Introduction}
\IEEEPARstart{L}{ight} detection and ranging (LiDAR) based \ac{PR} is critical for the robot navigation, multi-robot mapping \cite{huang2021disco, kangunified}, and \ac{SLAM} problems.
Unlike vision-based \ac{PR} methods \cite{arandjelovic2016netvlad, kim2023robust, choi2024referee} that have problems with illumination variations, visual changes, and time consumption in estimating 3-D information, LiDAR \ac{PR} approaches \cite{rohling2015fast, he2016m2dp, cop2018delight, kim2018scan, vidanapathirana2022logg3d, ma2022overlaptransformer, xu2023ring++} have become more prominent. % uy2018pointnetvlad, cao2020season, wang2020lidar, komorowski2021minkloc3d, zywanowski2021minkloc3d, yuan2023std, gupta2024effectively, yuan2024btc, shan2021robust

Although recent studies have shown remarkable results, reliable localization remains challenging in various situations. 
First, we often encounter situations where the \ac{FOV} is restricted through fusion with other sensors \cite{shin2020dvl} or occluded by mechanical components \cite{jeong2024diter} or the robot/sensor operator \cite{kim2020mulran}. 
Also, existing methods are challenging to deal with various vertical resolutions via LiDAR's channel option (e.g. 6 - 128) or different scanning patterns such as solid-state LiDAR \cite{lee2023conpr, jung2023helipr}. 
Moreover, we need to consider what information we can convey in the description (e.g. semi-metric) and processing time (i.e. descriptor generation and loop searching) as well as summarizing the place compactly.
% Moreover, conventional methods prioritize the performance and summarizing place compactly, rarely considering processing time (i.e. descriptor generation and loop detection).
Otherwise, traditional descriptors are challenging to be utilized on onboard computers. % \cite{8046794, 9561638}

\begin{figure}[t]
	\centering
	\def\width{0.48\textwidth}%
    {%
		\includegraphics[clip, trim= 10 10 5 5, width=\width]{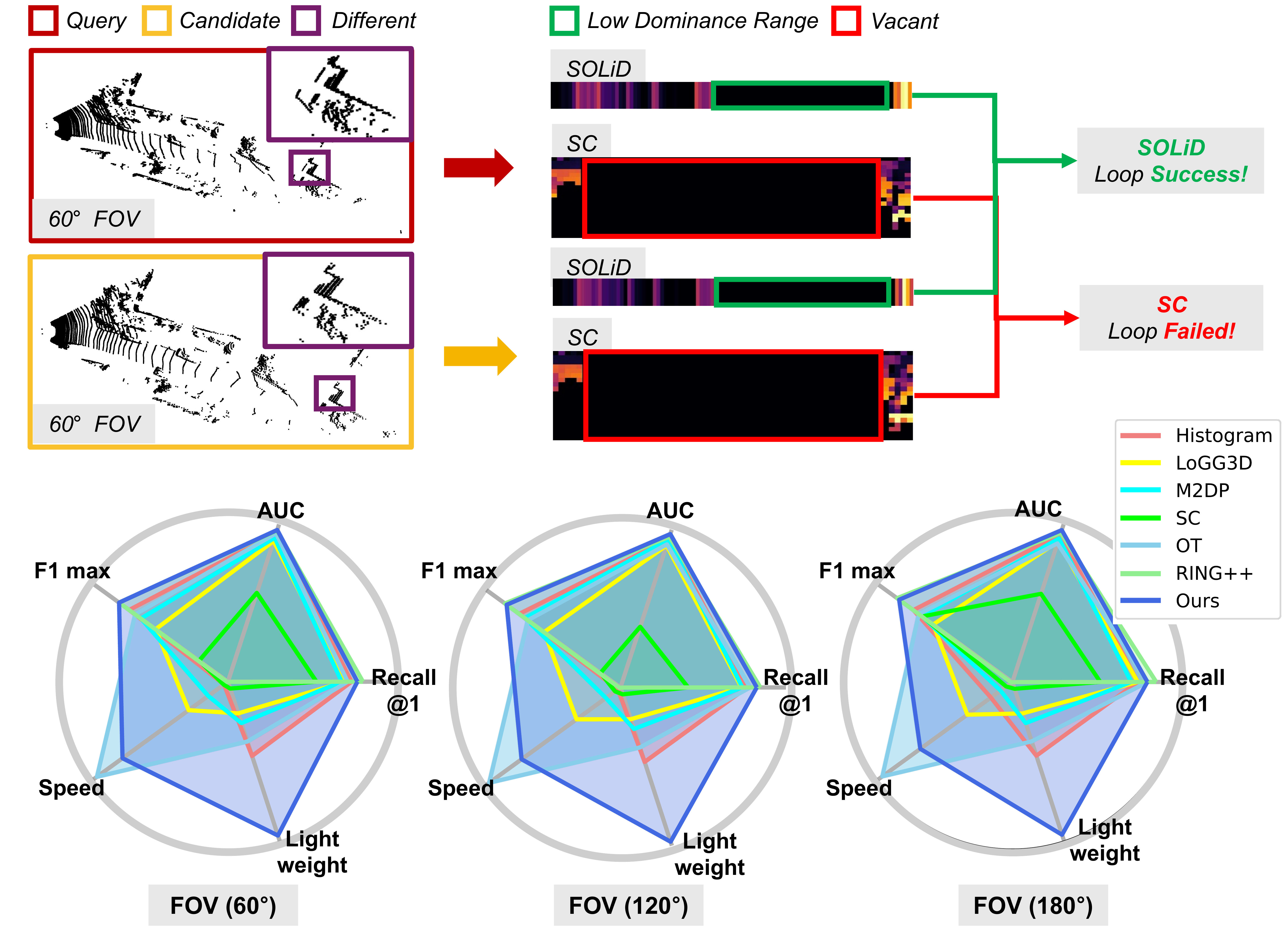}
	}
	\caption{At the top is our loop matching between the 1631st (red) and 192nd (yellow) frame in the KITTI 00 datasets clipped with a 60$^\circ$ FOV.
                 The loop matching exhibits slight differences in rotation.
                 % Our methodology (green) implies low points space, however, Scan Context (SC) (red) \cite{kim2018scan} signifies vacant information in the description.
                 Our methodology (green) implies a low dominance range, however, Scan Context (SC) (red) \cite{kim2018scan} signifies vacant information in the description.
                 As the FOV narrows, it becomes more sensitive to subtle changes in small (purple) areas, directly correlating with success or failure.
                 At the bottom are 3 phi charts that consist of recall@1, auc score, f1 max score, time, and descriptor size in KITTI 00 datasets with 60$^\circ$, 120$^\circ$, and 180$^\circ$ FOV.
                 Recall@1, auc score, and f1 max score are higher values, the better performance. 
                 The higher values of speed and lightweight represent the low computational cost and the lightweight descriptor.                 
                 We compare our method with well-known methods in LiDAR \ac{PR} \cite{rohling2015fast, he2016m2dp, kim2018scan, vidanapathirana2022logg3d, ma2022overlaptransformer, xu2023ring++} where discussed in Section \uppercase\expandafter{\romannumeral2}-A.
                             % At the bottom is the weight-to-area under the curve (AUC) score graph.
             % As the descriptor size increases, a more comprehensive description that effectively represents the scene is generated and the loop searching time is extended. 
             % In essence, a trade-off comes into play, and the effectiveness of this graph signifies how well it can overcome this limitation.
             % As shown in \tabref{table:eval_pr}, our method achieves higher scores within the red dotted square than RING++.
             % Especially, our performance is about 2×\times ∼\sim 1000×\times superior to Histogram ∼\sim State Of The Art (SOTA)~RING++ in the AUC score per descriptor size.
             }
        \vspace{-0.5cm}
	\label{fig:main}
\end{figure}

Unlike other approaches, we propose a spatially organized and lightweight global descriptor called \textit{SOLiD} for LiDAR \ac{PR} to overcome these constraints. 
The main contributions of this work are the following:
\begin{itemize}
    \item \textbf{Flexibility in Narrow FOV:} Our method applies to various localization scenarios with narrow FOV LiDAR as shown in \figref{fig:main}.
    Additionally, it ensures performance regardless of vertical resolution and can be applied to solid-state LiDAR with narrow FOV and different scan patterns.
    
    \item \textbf{Lightweight and Fast Description:} The proposed spatial encapsulation ensures lightweight description and fast searching. 
    As shown in \figref{fig:main}, our descriptor's performance is superior overall, making it well-suited for deployment on onboard computers with limited memory. 
    % As shown in \figref{fig:main}, our descriptor offers not only lightweight and fast speed but also performance, making it well-suited for deployment on onboard computers with limited memory. 
    Also, our method runs in real-time up to 80Hz without using GPU.
    
    \item \textbf{Rotational Robustness and Semi-metric Localization:} The proposed method achieves rotational robustness that can deal with rotation changes in narrow FOV.
    Furthermore, our approach provides a loop detection index and directly derives the 1-DoF angle (initial heading) to contribute to the accuracy of \ac{SLAM}.
    
    \item \textbf{Evaluation and Open-source: } We validate the proposed method in various challenging LiDAR localization scenarios (single session, multi-session, and multi-robots) with extensive quantitative metrics on localization performance and efficiency. 
    Also, the source code will be available and easy to utilize for existing SLAM frameworks \texttt{(\url{https://github.com/sparolab/solid.git})}. 
\end{itemize}

%% file: PaperWriting/relatedwork.tex
\section{Related Works}
LiDAR \ac{PR} methods can be broadly categorized into those that directly utilize geometric information (handcrafted methods) and those that leverage embedding vectors (deep learning methods).
To cope with various situations, LiDAR \ac{PR} systems must exhibit flexibility to handle limited FOV effectively.
Thus, we discuss traditional LiDAR \ac{PR} methodologies and their applicability to limited FOV situations.

\subsection{Place Recognition for LiDAR}
Histogram \cite{rohling2015fast} was crafted as a vector that gauges the intervals between points and constructed a cumulative distribution function based on the counts of these intervals.
M2DP \cite{he2016m2dp} analyzed the principal components of points in the current scene and projected the points onto these principal components to reduce dimensionality. 
Although these methods are lightweight 1-D descriptors, they require prolonged construction times. 
While pre-processing, such as downsampling, is possible, there is a trade-off with performance degradation, necessitating further consideration.
DeLight \cite{cop2018delight} generated 16 non-overlapping 3-D bins in spherical coordinates and incorporated intensity information into each bin.
Because the intensity is a relative value, intensity calibration is necessary.
LoGG3D-Net (LoGG3D) \cite{vidanapathirana2022logg3d} employed local consistency loss and local features to improve the performance of global descriptors.
OverlapTransformer (OT) \cite{ma2022overlaptransformer} based on range images utilized the transformer model to enhance the LiDAR \ac{PR} performance.
As with most deep learning networks, there is a generalization issue.
In addition, none of these methods proceed with estimating the initial pose.
SC \cite{kim2018scan} created 2-D bins in polar coordinates and designated the representative value of each bin as the height of the highest point.
Estimated rotation via column shifting could assist in pose correction using the \ac{ICP}.
RING \cite{lu2022one} was generated by the Radon Transform (RT) of a binary bird's eye view (BEV) image. 
Then, a translational invariant descriptor called TI-RING was created using the \ac{FFT} of RING.
However, as they are 2-D descriptors, memory efficiency diminishes over long-term autonomy.
RING++ \cite{xu2023ring++}, the SOTA method, generated 6 BEV images through 6 features similar to RING. 
However, RING++ is a 3-D descriptor, and the loop searching speed decreases more than the 2-D method.
% \begin{figure}[t]
% 	\centering
% 	\def\width{0.48\textwidth}%
%     {%
% 		\includegraphics[clip, trim= 70 570 70 10, width=\width]{Figure/limited_fov/limited_fov.pdf}
% 	}
% 	\caption{A variety of LiDAR tasks focused on specific FOV associated with Section \uppercase\expandafter{\romannumeral2}-B. 
%              Fig2. (a) presents the \textit{KITTI} object benchmark dataset \cite{geiger2012we}, providing approximately 90$^\circ$ of the front view. 
%              From left to right, the images depict raw data, \textit{PointSeg} \cite{wang2018pointseg}, and \textit{SqueezeSeg} \cite{wu2018squeezeseg}.
%              In Fig2. (b), DVL SLAM \cite{shin2020dvl} employs 360$^\circ$ LiDAR but utilizes only the overlapping points with images through camera-LiDAR fusion.
%              In Fig2. (c), MulRan dataset \cite{kim2020mulran}, the presence of radar overlap restricts the LiDAR's ability to observe the rear view comprehensively.}
%     \vspace{-0.5cm}
% 	\label{fig:limited_fov}
% \end{figure}

\subsection{Place Recognition in Limited FOV}
We often encounter limited FOV in robot navigation.
When the FOV is limited, the reduced number of bins in the constructed Histogram makes it challenging to describe the current scene effectively.
Moreover, M2DP exhibits performance limitations due to the constrained range of principal components, and LoGG3D's performance is also degraded due to the loss of the number of local features.
Additionally, when the orientation exceeds 180$^\circ$, 50$\%$ of the range image becomes vacant space, causing a loss of functionality in OT.
Furthermore, DeLight's 8 bins are transformed into empty space.
Similarly, SC becomes disabled as the description itself contains more than 50$\%$ blank space.
RING and RING++ benefit from the construction of rays from the RT, providing a relatively unrestricted capability in dealing with vacant spaces.
However, beneath this lies an inherent limitation for real-time performance, as the focus has been primarily on optimizing overall performance.

When considering various methodologies, it is observed that traditional approaches including \cite{kim2018scan, xu2021disco, lu2022one, xu2023ring++, ma2023cvtnet} often rely on the BEV representation, which projects the z-axis. 
However, in scenarios where the FOV is restricted, LiDAR's horizontal and vertical angles are similar, and in some cases, the vertical angle can even be significant such as Livox AVIA (70.4$^\circ$ (Horizontal) × 77.2$^\circ$ (Vertical)).
% While techniques utilizing a range image \cite{shan2021robust, ma2022overlaptransformer} exist, they often lack rotational invariance, leading to failures in scenarios such as reverse loops.
% Therefore, drawing inspiration from both the translational invariant range image and rotational invariant image projected onto the angle axis, we describe the place efficiently and compactly.

% \subsection{Place Recognition with Limited FOV}
% LiDAR \ac{PR} methods focus on leveraging a complete 360∘^\circ pointcloud.
% However, in practical scenarios, especially for enhancing autonomous driving, the integration of deep learning for object detection \cite{wang2018pointseg, wu2018squeezeseg, wu2019squeezesegv2} and road segmentation \cite{aksoy2020salsanet} with LiDAR often involves camera-LiDAR fusion \ac{SLAM} \cite{shin2020dvl}. 
% This fusion typically utilizes only the overlapping LiDAR views with camera information, negating the need to process all 360∘^\circ data, as shown in \figref{fig:limited_fov}.
% In addition, LiDAR failures that lead to occlusions unavoidably restrict the use of its 360$^\circ$ capability. 
% In addition, there has been an increase in the use of solid-state LiDARs and public datasets \cite{kim2020mulran, lee2023conpr, jung2023helipr} that focus on front views.
% This trend signifies the successful implementation of LiDAR odometry algorithms with a limited FOV \cite{lin2020loam, xu2021fast, li2021towards, wang2022tightly}, as evidenced by their applications in the delivery robot market \cite{9981281}.

%% file: PaperWriting/method.tex
\section{Method}
% The LiDAR \ac{PR} can be broadly categorized into three stages: place description, which explains the current location; loop detection, which identifies revisited places based on the description; and pose estimation, which determines the relative pose between the current location and revisited places.
% In LiDAR \ac{PR}, as the description becomes more captured, performance can only be expected to improve.
% However, in terms of complexity, portraying a scene lies in the hidden inefficiency of time-consuming loop searching.
% Similarly, as the scene becomes more compressed, the performance decreases, but this allows for rapid search times.
% Therefore, a spatially organized and lightweight global description as represented in \figref{fig:pipeline} is necessary to encapsulate the current place effectively.
% Next, the two-step search process is introduced. The overall pipeline of place recognition using scan context is depicted in Fig. 2. 
% The Scan Context creation and validation can also be found in 
% and propose a measure that calculates the distance between two \textit{SOLiD}.

\begin{figure*}[t]
	\centering
	\def\width{0.99\textwidth}%
        {%
		\includegraphics[clip, trim= 0 190 0 160, width=\width]{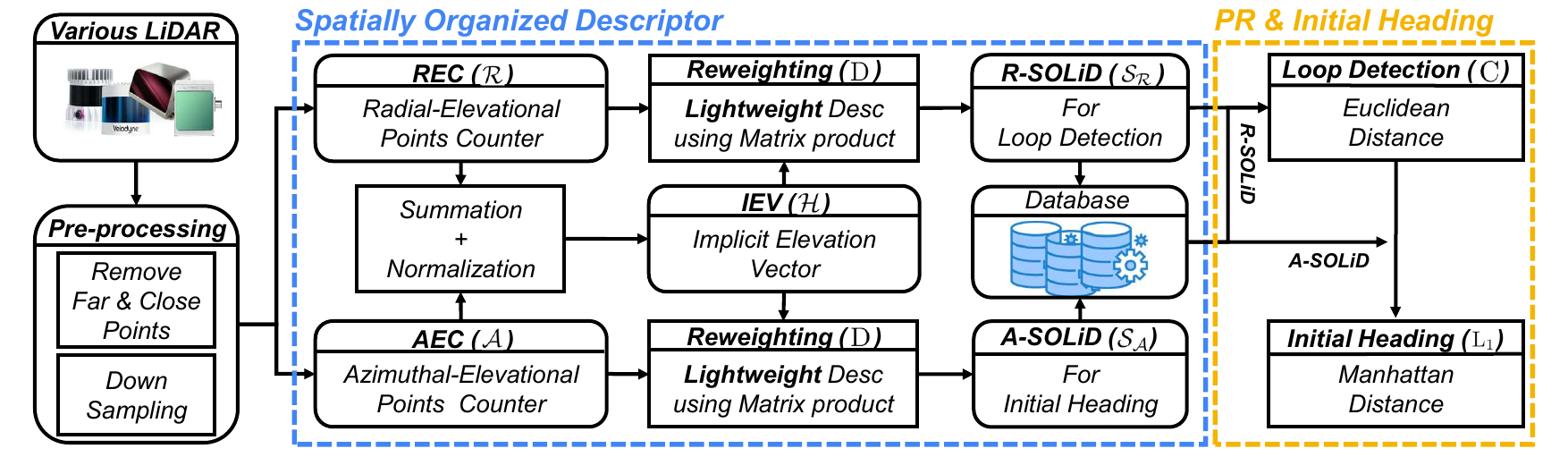}
	}
    \vspace{-0.8cm}
     \caption{Overall pipeline of our algorithm. Angled and rounded rectangles represent data and algorithms.
	}
    \vspace{-0.6cm}
    \label{fig:pipeline}
\end{figure*}

% spatial organization의 정의를 어떻게 할 수 있을까?
In this section, we define the spatial organization and describe \textit{SOLiD} generation from a 3-D scan and calculation of the distance between two \textit{SOLiD} as represented in \figref{fig:pipeline}.
% Spatial Organization is ...
Spatial organization refers to organizing a 3-D space into 2-D bins along the range-elevation and azimuth-elevation directions using polar coordinates and determining representative values for these bins.
Unlike the BEV representation, the spatial organization includes elevation information, allowing for multi-directional analysis of 3-D space.
Therefore, to achieve place recognition within a limited FOV, we utilize spatial organization and create \textit{SOLiD} by reweighting vertical information.

% To achieve spatial organization in limited FOV, we propose a representation combining vertical, radial, and azimuthal information.
% Next, we create a spatially organized and lightweight global descriptor for FOV-constrained LiDAR place recognition by reweighting the vertical information.

% ======================================================================================================================
%% PipeLine Figure
% ======================================================================================================================

\subsection{Spatially Organized Place Description}
\subsubsection{REC ($\mathcal{R}$) $\&$ AEC ($\mathcal{A}$)}
% We define a place descriptor called \textit{SOLiD} for FOV-constrained LiDAR PR.
The key idea of \textit{SOLiD} is inspired by the counting algorithm \cite{rohling2015fast} and spherical representation \cite{cop2018delight}. 
First, we define the 3-D space in spherical coordinates.
Second, we organize the 3-D space into 3-D bins in the radial, azimuthal, and elevational directions. 
Finally, we downsample the points to a voxel size of 0.5m and count the number of points within each 3-D bin.
% Specifically, we define a 3-D space in spherical coordinates and divide it into 3-D bins.
% As shown in \figref{fig:method}, 
% If points $x$, $y$, and $z$ of the current 3-D scan, we convert it into spherical coordinates as below:
The points $x$, $y$, and $z$ in the current 3D scan are converted to spherical coordinates as below:
\begin{equation}
    \begin{cases}
    r      = \sqrt{x^2 + y^2}     \\ 
    \theta = \text{atan2}(y, x)          \\ 
    \phi   = \text{arctan}(\frac{z}{r}) .
    \end{cases}
    \label{equation:spherical_coordinates}
\end{equation}

\begin{figure}[t]
	\centering
	\def\width{0.48\textwidth}%
        {%
		\includegraphics[clip, trim= 0 0 0 80, width=\width]{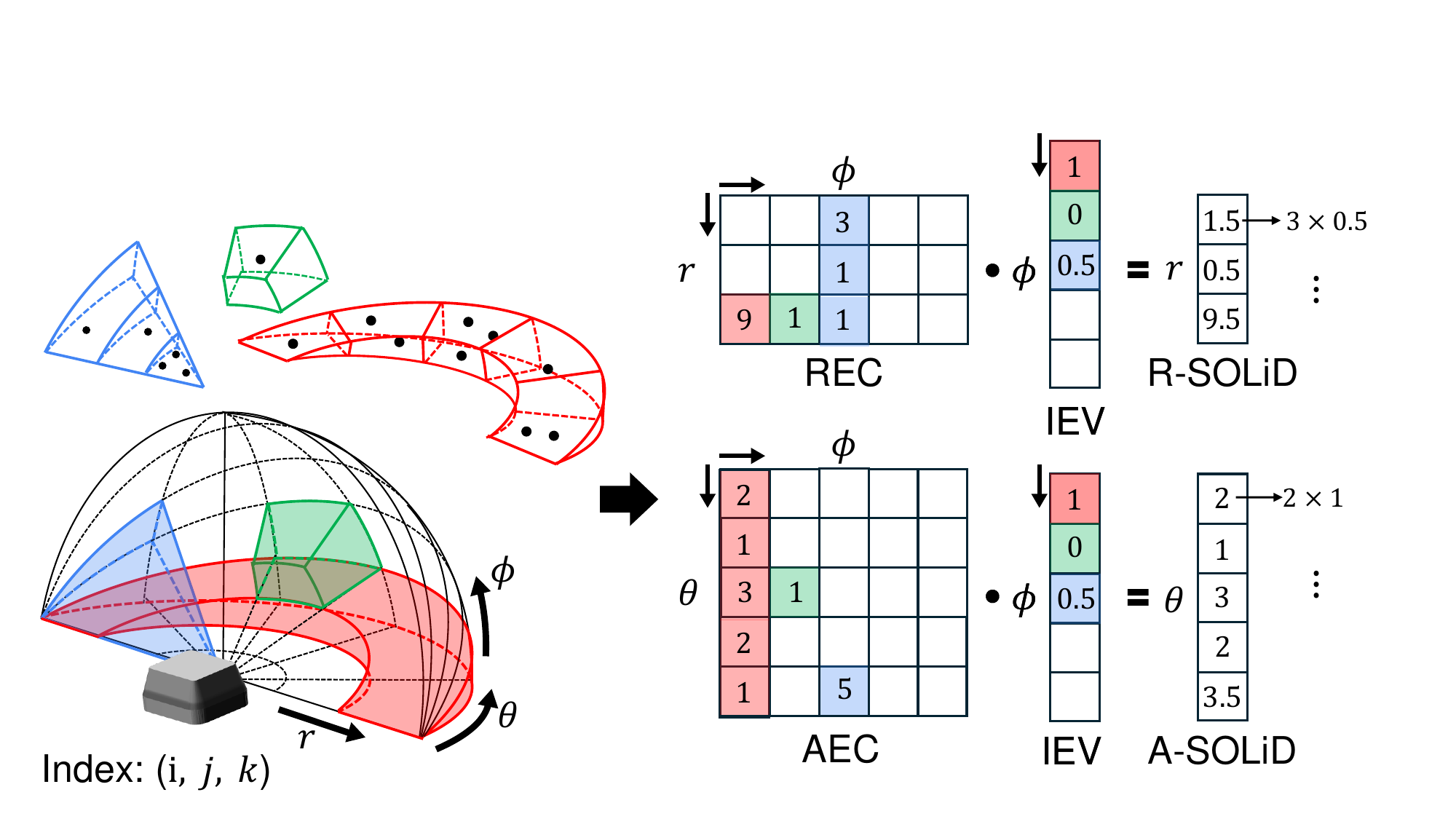}
	}
    \vspace{-0.6cm}
    \caption{Simple example schema of our methodology.
             The indices $i$, $j$, $k$ are determined along the $r$, $\theta$, $\phi$ axes.
             Utilizing the spatial organization, 3-D bins are projected onto \textit{REC} and \textit{AEC} according to their respective indices. 
             Subsequently, \textit{R-SOLiD} and \textit{A-SOLiD} are generated by taking the matrix product with \textit{IEV}.
             }
    \label{fig:generation}
    \vspace{-0.65cm}
\end{figure}

% \subsubsection{REC (R\mathcal{R}) &\& AEC (A\mathcal{A})}
\noindent The number of points in each bin signifies the importance of the 3-D bin itself.
Therefore, we assign each bin the number of points as a representative.
However, we bear the burden of high computational costs owing to the heavyweight size if we utilize the 3-D bins as a place descriptor.
Also, these bins often suffer from the issue of being empty, and the sparsity increases with distance.
Unlike existing methods using elevational projection (i.e. BEV) \cite{kim2018scan, xu2021disco, lu2022one, xu2023ring++, ma2023cvtnet}, we utilize the spatial organization in both the radial and azimuthal directions to generate dense representations, namely, \textit{AEC} ($\mathcal{A}$) and \textit{REC} ($\mathcal{R}$), as follows: % project the 3-D scan -> utilize the spatial organization
\begin{equation}
    \mathcal{R} = \bigcup_{i\in [N_r], \; k\in [N_e]} \mathcal{R}_{ik}, \; \; \; \; \; \;
    \mathcal{A} = \bigcup_{j\in [N_a], \; k\in [N_e]} \mathcal{A}_{jk},
    \label{equation:REC}
\end{equation}
% \begin{equation}
%     \mathcal{A} = \bigcup_{j\in [N_a], \; k\in [N_e]} \mathcal{A}_{jk},
%     \label{equation:AEC}
% \end{equation}

\noindent where $N_r$, $N_a$, and $N_e$ denote the number of radial, azimuthal, and elevational bins, respectively. 
The symbols $[N_r]$, $[N_a]$, and $[N_e]$ are equal to $\{ 1, 2, ... ,N_{r-1}, N_r\}$, $\{ 1, 2, ... ,N_{a-1}, N_a\}$, and $\{ 1, 2, ... ,N_{e-1}, N_e\}$.
Each $\mathcal{R}$ and $\mathcal{A}$ is composed of indices $i$, $k$ and $j$, $k$, respectively, and indices $i$, $j$, $k$ ($i, j, k \in \mathbb{Z}$) can be determined by range $r$, azimuth $\theta$, elevation $\phi$ as follows:

\begin{equation}
    i \in \left\{ \mathrm{R} \bigg( N_r \times \frac{r}{L_{max}}+1 \bigg) \biggm| \; 0          \leq r      < L_{max}, \; r \in \mathbb{R}   \right\},   \\ 
    \label{equation:index}
\end{equation}
\begin{equation}
    j \in \left\{ \mathrm{R} \bigg( N_a \times \frac{\theta}{360}+1 \bigg) \biggm| \; 0          \leq \theta < 360,  \; \theta \in \mathbb{R}      \right\},  \\ 
    \label{equation:index}
\end{equation}
\begin{equation}
    k \in \left\{ \mathrm{R} \bigg( N_e \times \frac{\phi - F_{d}}{F_{u} - F_{d}}+1 \bigg) \biggm| \; F_{d} \leq \phi < F_{u}, \; \phi \in \mathbb{R}  \right\},
    \label{equation:index}
\end{equation}

\noindent where function $\mathrm{R}(\cdot,\cdot)$ means round function, $L_{max}$ is the maximum observable distance and, $F_{u}$ and $F_{d}$ represent the down and up vertical FOV of the LiDAR sensor, respectively. 
As shown in \figref{fig:generation}, the number of points in the index $i$, $j$, $k$ bin become the values that make up the \textit{REC} and \textit{AEC}.
As a result, we do not overlook vertical information by describing a place using spatial organization within a limited FOV where the amount of information in the vertical direction is comparable to that in the horizontal direction.

% \begin{figure}[t]
% 	\centering
% 	\def\width{0.45\textwidth}%
%     {%
% 		\includegraphics[clip, trim= 0 0 0 0, width=\width]{Figure/method/method.pdf}
% 	}
%         \vspace{-0.3cm}
%  \caption{The process of generating \textit{IEV}. \textit{EC} is created through elevation-wise summation of \textit{REC}. Then, \textit{EC} is adapted to min-max normalization to produce \textit{IEV}.
%             }
%         \vspace{-0.3cm}
% 	\label{fig:method}
% \end{figure}

\begin{figure}[t]
	\centering
	\def\width{0.45\textwidth}%
    {%
		\includegraphics[clip, trim= 130 300 150 50, width=\width]{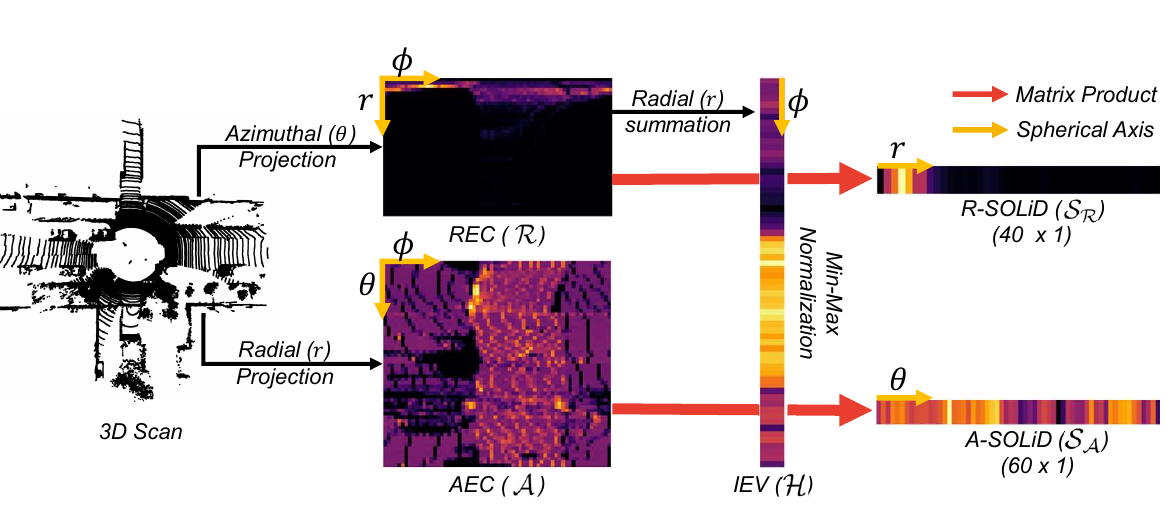}
	}
        \vspace{-0.3cm}
 \caption{The process of generating \textit{R-SOLiD} and \textit{A-SOLiD}. 
         First, when a 3-D scan is received, \textit{REC} and \textit{AEC} are generated through spatial organization. 
         Second, summation is performed in the radial direction, followed by the creation of \textit{EC}. 
         Third, min-max normalization is applied to the created \textit{EC}, resulting in the generation of \textit{IEV}. 
         Finally, the matrix product of \textit{IEV} with \textit{REC} and \textit{AEC} respectively yields \textit{R-SOLiD} and \textit{A-SOLiD}.
            }
        \vspace{-0.65cm}
	\label{fig:method}
\end{figure}

% In addition, the angles in the vertical direction tend to be similar to those in the horizontal direction.

% \subsubsection{IEV ($\mathcal{H}$)}
\subsubsection{IEV ($\mathcal{H})$}
% Relying only on \textit{REC} or \textit{AEC} renders the system vulnerable to issues. % such as noise and distortion.
% For instance, in a solid-state LiDAR, even for the same place, the number of points may vary because of non-repetitive scan patterns.
Using only \textit{REC} or \textit{AEC} for place recognition problems can cause system performance degradation. 
For instance, on a solid-state LiDAR, the number of points belonging to a 3-D bin can be different due to non-repetitive scan patterns, even at the same place.
Also, in limited FOV situations, \textit{REC} information becomes uncertain, and \textit{AEC}, similar to SC, incorporates vacant spaces as described in \figref{fig:main}.
To address this issue, we apply reweighting using \textit{IEV} ($\mathcal{H}$), which effectively highlights the spatially dominant components.
To obtain \textit{IEV}, we first create \textit{EC} by summing \textit{REC} in the radial direction, as represented in \figref{fig:method}.
Summing \textit{AEC} in the azimuthal direction also becomes \textit{EC}, but since $N_r$ is typically smaller than $N_a$, generating \textit{EC} via \textit{REC} is more time-efficient.
Thus, the \textit{EC} is represented as follows:
\begin{equation}
    \mathcal{E} = [{\psi}_1, {\psi}_2, ..., {\psi}_{N_e}], \qquad \psi_k = \sum_{i=1}^{N_r} \mathcal{R}_{ik}.
    \label{equation:IEV}
\end{equation}
\textit{EC} is a measure that examines which vertical angles have a higher number of points (i.e. dominant) in an elevation-wise manner.
However, as mentioned above, the number of points is vulnerable to various issues such as non-repetitive scan patterns.
Therefore, we generate \textit{IEV} ($\mathcal{H}$) through min-max normalization of \textit{EC} ($\mathcal{E}$) to identify the tendency for vertical structure at the current place as below:
\begin{equation}
    \mathcal{H} = \frac{\mathcal{E}-\mathcal{E}_{min}}{\mathcal{E}_{max}-\mathcal{E}_{min}},
    \label{equation:IEV}
\end{equation}
where $\mathcal{E}_{min}$ and $\mathcal{E}_{max}$ are min and max value in \textit{EC} ($\mathcal{E}$).
As a result, \textit{IEV} is a robust weight that includes elevation-wise dominance and can cope with different scan patterns.

\subsubsection{SOLiD ($\mathcal{S}$)}
Finally, we acquire a \textit{SOLiD} by reweighting, which takes the matrix product between \textit{REC}, \textit{AEC}, and \textit{IEV} ($\mathcal{H}$) are as follows:
\begin{equation}
    \mathcal{S} = [\mathcal{S}_\mathcal{R}, \mathcal{S}_\mathcal{A}],
    \qquad
    \begin{cases}
    \mathcal{S}_\mathcal{R}      = \mathrm{D}(\mathcal{R} , \mathcal{H})   \\ 
    \mathcal{S}_\mathcal{A}      = \mathrm{D}(\mathcal{A} , \mathcal{H}),  \\ 
    \end{cases}
    \label{equation:solid}
\end{equation}
where function $\mathrm{D}(\cdot,\cdot)$ means matrix production.
\textit{R-SOLiD} ($\mathcal{S}_\mathcal{R}$) and \textit{A-SOLiD} ($\mathcal{S}_\mathcal{A}$) form vectors that encapsulate radial and azimuthal dominance.
This strategy propagates the current place's element-wise dominance to each range and azimuth.
In detail, the element-wise multiplication and summation are performed between all elevation elements within each range and azimuth of \textit{REC}, \textit{AEC}, and all elements of \textit{IEV} as represented in \figref{fig:generation}.
This step is called reweighting, where higher-value elements converge higher-value elements than before and lower-value elements converge lower-value than before.
Each \textit{SOLiD} is a vector representing the degree of dominance for the given range and azimuth within the current place.
In this work, $\mathcal{S}_{\mathcal{R}}$ is used for loop detection, and $\mathcal{S}_{\mathcal{A}}$ is used for initial heading estimation, where each consists of sizes $N_r$ and $N_a$.
In addition, we determine that $N_r = 40$, $N_a = 60$, and $N_e$ are aligned with the vertical resolution of the LiDAR.

\subsection{Place Recognition \& Initial Heading Estimation}
\subsubsection{Place Recognition} 
To recognize the revisited place, we compare the cosine distance between the query \textit{R-SOLiD} ($\mathcal{S}_{\mathcal{R}}^q$) and the candidate \textit{R-SOLiD} ($\mathcal{S}_{\mathcal{R}}^c$) in the database as follows:
\begin{equation}
    \mathrm{C}(\mathcal{S}_{\mathcal{R}}^q,\mathcal{S}_{\mathcal{R}}^c) = \left(1 - {{\mathcal{S}_{\mathcal{R}}^q \cdot \mathcal{S}_{\mathcal{R}}^c} \over ||\mathcal{S}_{\mathcal{R}}^q||\cdot ||\mathcal{S}_{\mathcal{R}}^c||}\right).
    \label{equation:cossim}
\end{equation}
\noindent By constructing a kd-tree, we search for candidates with the closest distances in the database and recognize them as revisited places. 
\subsubsection{Initial Heading Estimation}
Owing to the semi-metric representation of the $\mathcal{S}_{\mathcal{A}}$, each unit movement in the vector $\mathcal{S}_{\mathcal{A}}$ represents a corresponding heading (yaw).
Thus, we find the nearest distance index $n^*$, shifting the candidate \textit{A-SOLiD} ($\mathcal{S}_{\mathcal{A}_{n}}^c$) by $n$ unit as follows:
\begin{equation}
    n^* = \argmin\limits_{n \in [N_a]}({||\mathcal{S}_{\mathcal{A}}^q - \mathcal{S}_{\mathcal{A}_n}^c||}) .
    \label{equation:initial_index}
\end{equation}

%% file: PaperWriting/experiments2.tex
\section{Results}
% 1) 어떤 방법론과 비교 했고, 어떤 cpu/gpu에서 했다. 
% 2) Scenarios and Metrics
%  - single / multi-session / multi-robot 시나리오 설명
%   : single 에서는 well-known kitti를 활용해 다양한 narrow fov에 대한 성능과 LiDAR scan occlusion에 대한 검증
%   : multi-session에서는 Narrow FOV LiDAR와 Solid-state LiDAR에 대해서 검증
%   : Multi-robot은 communication을 가정하고 communication time에 대해서 검증
%  - evaluation metric 어떻게 했는지 설명

% 3) processing time에 대해서는 따로 뺄 수 있나요? 가능하다면 description time과 searching time을 구분해두면 좋겠습니다. 또한, kd-tree가 들어간 searching time도 보여주면 좋구요. 
All experiments were conducted on desktop Intel i7-12700 KF with RTX 3070Ti and onboard computer Jetson AGX Orin to assess onboard computing capabilities.
Furthermore, we differentiated experiment results for easy comprehension: \bl{\textbf{first rank}} in bold blue, the \gl{\textbf{second rank}} in green bold, and the \textbf{third rank} in black bold, respectively.

\subsection{Experiments Setup}
% ===================================================================================================================================================================================
\subsubsection{Comparative Methods}
We presented Histogram \cite{rohling2015fast}, M2DP \cite{he2016m2dp}, SC \cite{kim2018scan}, LoGG3D \cite{vidanapathirana2022logg3d}, OT \cite{ma2022overlaptransformer}, and RING++ \cite{xu2023ring++} as comparative methods, generated using Python-based code to meet the same conditions.
We also differentiated between methods that can be used only with CPU and those where GPU usage is inevitable.
Each method was labeled for easy identification as either (C) for CPU-only or (G) for GPU-dependent.
Especially, we utilized the pre-trained models provided for deep-learning methods (LoGG3D and OT).

%%% 여기부터 해야함.
Each of these methods employs different search strategies.
Among them, radius search strongly assumes that the initial pose is known.
Therefore, we converted radius search into brute force searching.
For a fair comparison, the proposed method also derived results exclusively using brute force searching.
The results obtained using the kd-tree method were explained in Section \uppercase\expandafter{\romannumeral4}-D.

% ===================================================================================================================================================================================

% ===================================================================================================================================================================================
\subsubsection{Scenarios}
\begin{figure}[h]
    \centering
    \def\width{0.98\columnwidth}
    \includegraphics[clip, trim= 20 220 20 180, width=\width]{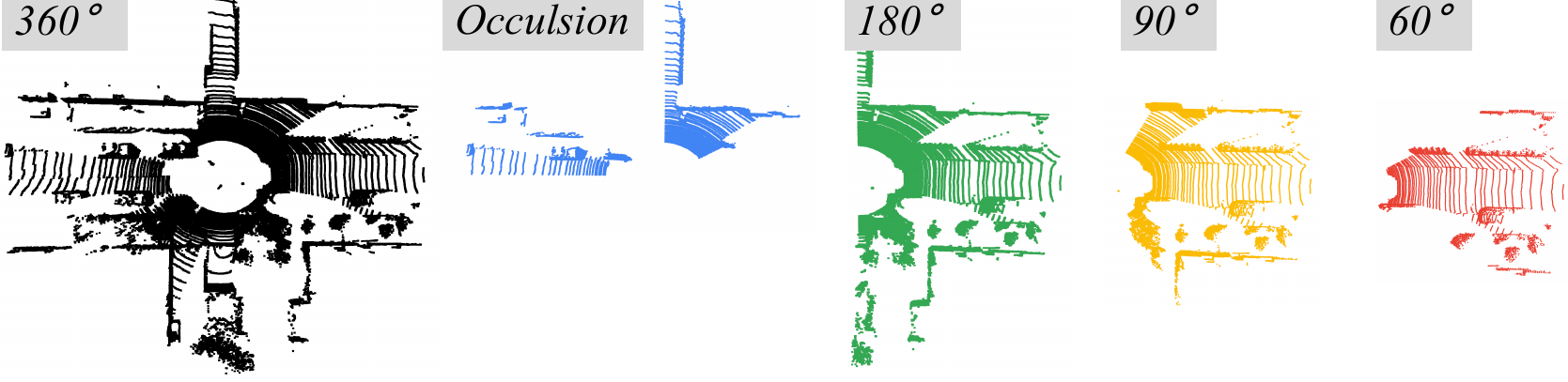}
    \caption{500th frame from various angles in KITTI 00. Black represents full view (360$^\circ$) points. Blue, green, yellow, and red mean occlusion, 180$^\circ$, 120$^\circ$, and 60$^\circ$ points, respectively.}
    \label{fig:various_fov}
    \vspace{-0.2cm}
\end{figure}

For extensive evaluation, we validated across various scenarios (single session, multi-session, and multi-robot system).
For each session, we deployed different LiDARs (different vertical angles, horizontal angles, channels, or scan patterns).
As shown in \figref{fig:various_fov}, we modified 360$^\circ$ FOV datasets to restricted FOV datasets by clipping the pointcloud to assess performance in dealing with limited FOV.

We validated the performance across various narrow FOV scenarios and LiDAR scan occlusion situations in the well-known KITTI dataset \cite{geiger2012we} of a single session. %HDL-64E
KITTI was a sequence of driving cars equipped with Velodyne HDL-64E LiDAR. 
We evaluated various metrics through loops within 10 meters using sequences 00, 02, and 05.

In a multi-session HeLiPR dataset \cite{jung2023helipr}, we evaluated the performance of solid-state LiDAR with heterogeneous scan patterns and limited FOV. % Aeva Avia
HeLiPR was a sequence of driving cars, and we tested the robustness of PR through solid-state LiDARs, Aeva Aeries \uppercase\expandafter{\romannumeral2} and Livox-Avia.
In this experiment, we leveraged the sequences KAIST04-05, Town01-02, and RoundAbout01-02. 
In this dataset, the vehicle remained in the same spot for extended periods due to waiting at traffic lights.
Therefore, we sampled the pointcloud at intervals of 2 meters based on the ground truth pose and validated using loops within 5 meters.

In a multi-robot scenario Park dataset \cite{huang2021disco}, we measured the performance across various constrained FOV setups and captured the communication time between robots. %VLP-16
Park was a sequence where three robots equipped with Velodyne VLP-16 LiDAR map a designated area.
Since each robot needs to be aware of its relationship with the other two robots, we evaluated PR based on several metrics through six experiments.
Since robot driving was inevitably slower than driving cars, we sampled at intervals of 2 meters and considered loops within 5 meters as correct loops.
% ===================================================================================================================================================================================
\vspace{-0.5cm}
\subsection{Evaluation Metrics}
\subsubsection{PR curve}
For evaluating the performance of loop detection, the Precision-Recall (PR) curve was utilized. 
Precision and recall \cite{yin2024survey} are defined as below:
\begin{equation}
\text { Precision }=\frac{\mathrm{TP}}{\mathrm{TP}+\mathrm{FP}}, \quad \text { Recall }=\frac{\mathrm{TP}}{\mathrm{TP}+\mathrm{FN}},
\end{equation}
where TP, FP, and FN are true positive, false positive, and false negative, respectively. 

\subsubsection{ROC curve and AUC score}
We also plotted the Receiver operating characteristic (ROC) curve to assess the ability to distinguish between true and false loops.
True Positive Rate (TPR) and False Positive Rate (FPR) are defined as below:
\begin{equation}
\text { FPR }=\frac{\mathrm{FP}}{\mathrm{FP}+\mathrm{TN}}, \quad \text { TPR }=\frac{\mathrm{TP}}{\mathrm{TP}+\mathrm{FN}},
\end{equation}
where TP, FP, and TN, FN are true positive, false positive, and true negative, false negative, respectively. 
We also evaluated the AUC score, which represents the area under the ROC curve.
TPR should be greater than FPR for a model to diagnose true and false loops.
Thus, the method has no discriminatory ability if the AUC score is less than 0.5.

\subsubsection{Recall@1 and F1 score}
We derived Recall@1 as a metric for performance in finding loops, rather than the PR curve or ROC curve. % \cite{shi2023lidar} 
Additionally, we derived the F1 score as a metric for performance, which is the harmonic mean between Precision and Recall.
Each Recall@1 and F1 score \cite{yin2024survey} metric is defined as follows:
\begin{equation}
\text { Recall@1 }=\frac{\mathrm{TP}}{\mathrm{GT}}, \quad \text { F1 score}=\frac{2 \times \mathrm{Precision} \times \mathrm{Recall}}{\mathrm{Precision}+\mathrm{Recall}},
\end{equation}
where TP, GT, Precision, and Recall are true positive, the number of ground truth, precision, and recall are mentioned above, respectively. 

\subsubsection{Rotation Error}
We evaluated a Rotation Error (RE) \cite{yin2024survey}, which is the error between the estimated rotation and the ground truth rotation calculated by
\begin{equation}
\text { RE }= \text {arccos}((\text {trace}(\hat{\mathrm{R}}^T \cdot \mathrm{R}) - 1)/2),
\end{equation}
where $\hat{\mathrm{R}}^T$ is the estimated rotation matrix and $\mathrm{R}$ is the ground truth rotation matrix.

\subsubsection{Processing Time}
In PR experiments, much time is consumed in both the process of generating descriptions and finding loops.
The process of generating descriptions remains consistent even as the database size increases. 
However, loop searching time inevitably increases as the database grows.
Therefore, we designated and measured the processing time by summing the average time of the entire loop searching process with the time taken for description generation.
We verified the processing time on the desktop and onboard computer, labeling them respectively as (D) and (O).
Especially in the case of onboard computers (O), including laptops installed on real robots, GPU-based methods were not considered in ranking processing time due to the lack of GPU.
% ===================================================================================================================================================================================

% ===================================================================================================================================================================================
\subsection{Lightness}
As represented in \tabref{table:description}, we measured the size and shape of each descriptor for comparison.
The proposed method is at least approximately 2$\times$ lighter, up to 1000$\times$ lighter compared with traditional methods.
Thus, our method is more effective in large-scale mapping as it can be stored for longer periods in a finite number of storages.
Especially, our method is more effective on onboard computers with limited memory capacity.
In the following sections, we can see that our method performs comparatively or better than other methods.
This suggests that the proposed method is lightweight and achieves robust PR using spatial organization and reweighting in limited FOV.
In particular, RING++ employs a 3-D descriptor of size 6$\times$120$\times$120, as it minimizes compression in the 3-D space.
While this lack of dimension reduction allows for a more diverse representation of the scene, it is natively a heavyweight descriptor.
Furthermore, the processing time naturally increases because heavyweight descriptors have more space to compare.
\begin{table}[h]
\vspace{-0.3cm}
\caption{Comparison of Descriptor sizes}
\renewcommand{\arraystretch}{1.5}
\centering\resizebox{0.48\textwidth}{!}{
{\huge
\begin{tabular}{c|ccccccc}
\toprule \hline
Descriptor       & Hist. (C)          & LoGG3D (G)   & M2DP (C)     & SC (C)             & OT (G)        & RING++ (G)              & Ours (C) \\ \hline
Shape            & 100$\times$1       & 256$\times$1 & 192$\times$1 & 20$\times$60       & 256$\times$1  & 6$\times$120$\times$120 & 40$\times$1  \\ \hline
Size [B]         & \gl{\textbf{928}}  & 2176         & 1664         & 10016              & \textbf{1152} & 345728                  & \bl{\textbf{448}}  \\ \hline \bottomrule
% Initial Guess    & {x}                & {x}        & {x}          & \checkmark         & {x}           & \checkmark   & \checkmark    \\ \hline \bottomrule
\end{tabular}}}
\label{table:description}
\vspace{-0.4cm}
\end{table}
% ===================================================================================================================================================================================

% ===================================================================================================================================================================================
\subsection{Single session Place Recognition}
\subsubsection{Narrow FOV}
As described in \tabref{table:eval_pr}, our approach and RING++ outperform across various metrics than other approaches.
The proposed method finds fewer loops overall compared to RING++.
However, our method has an outstanding ability to distinguish between correct loops and incorrect loops, which affects SLAM performance.
Especially in terms of time, it is overwhelmingly superior.
Therefore, our method maintains a consistent map efficiently and quickly even in limited FOV.
%% ===========================================================================================================================================================================
% \begin{table}[h]
% \caption{Processing Time on Onboard Computer}
% \renewcommand{\arraystretch}{1.5}
% \centering\resizebox{0.48\textwidth}{!}{
% \begin{tabular}{c|ccccccc}
% \toprule \hline
% \multirow{2}{*}{FOV} & \multicolumn{7}{c}{Processing Time (O) {[}s{]}}                                     \\ \cline{2-8} 
%                      & Hist. (C) & LoGG3D (G)          & M2DP (C) & SC (C) & OT (G)               & RING++ (C) & Ours (C) \\ \hline
% 60                   & 1.289     & \textbf{0.160}      & 0.297    & 1.359  & \bl{\textbf{0.036}}  & 10.85      & \gl{\textbf{0.073}}    \\
% 120                  & 2.442     & \textbf{0.188}      & 0.531    & 1.371  & \bl{\textbf{0.047}}  & 10.92      & \gl{\textbf{0.093}}    \\
% 180                  & 3.550     & \textbf{0.215}      & 0.710    & 1.382  & \bl{\textbf{0.056}}  & 10.98      & \gl{\textbf{0.111}}    \\ \hline \bottomrule
%                                % & 360                  & 6.996     & \textbf{0.312}      & 1.136    & 1.440  & \bl{\textbf{0.083}}  & 11.16      & \gl{\textbf{0.200}}    \\ \hline \bottomrule
% \end{tabular}}
% \vspace{-0.3cm}
% \label{table:fov_time}
% \end{table}
% ===================================================================================================================================================================================
\begin{table*}[t]
\caption{Evaluation in KITTI Datasets}
\centering\resizebox{\textwidth}{!}{
\begin{tabular}{cccccccccccccc}
\toprule \hline
\multicolumn{1}{c|}{\multirow{3}{*}{FOV}} & \multicolumn{1}{c|}{\multirow{3}{*}{Method}} & \multicolumn{12}{c}{Sequence}                                                                                                                                                                                                                                                                                                                                                                                               \\ \cline{3-14}
\multicolumn{1}{c|}{}       & \multicolumn{1}{c|}{} & \multicolumn{4}{c|}{00 (4541 frames)}  & \multicolumn{4}{c|}{02 (4661 frames)} & \multicolumn{4}{c}{05 (2761 frames)}\\ \cline{3-14} 
\multicolumn{1}{c|}{}                          & \multicolumn{1}{c|}{}           & Recall@1                                & \multicolumn{1}{c}{AUC} & \multicolumn{1}{c}{F1 max} & \multicolumn{1}{c|}{Time (D) [Hz]}              
                                                                                 & \multicolumn{1}{c}{Recall@1}            & \multicolumn{1}{c}{AUC} & \multicolumn{1}{c}{F1 max} & \multicolumn{1}{l|}{Time (D) [Hz]} 
                                                                                 & \multicolumn{1}{c}{Recall@1}            & \multicolumn{1}{c}{AUC} & \multicolumn{1}{c}{F1 max} & Time (D) [Hz] \\ \hline
\multicolumn{1}{c|}{\multirow{7}{*}{60}}       & \multicolumn{1}{c|}{Hist. (C)}  & \multicolumn{1}{c}{0.764}               &0.965                    &0.754                       & \multicolumn{1}{c|}{3.06}          
                                                                                 & 0.483                                   &0.960                    &0.452                       & \multicolumn{1}{c|}{3.05}     
                                                                                 & 0.611                                   &0.949                    &0.606                       & 3.21     \\
\multicolumn{1}{c|}{}                          & \multicolumn{1}{c|}{Logg3D (G)} & \multicolumn{1}{c}{0.726}               &0.914                    &0.552                       & \multicolumn{1}{c|}{30.76}
                                                                                 & 0.674                                   &0.954                    &0.627                       & \multicolumn{1}{c|}{30.59}
                                                                                 & 0.593                                   &0.936                    &0.531                       & 33.55     \\
\multicolumn{1}{c|}{}                          & \multicolumn{1}{c|}{M2DP (C)}   & \multicolumn{1}{c}{0.703}               &0.935                    &0.664                       & \multicolumn{1}{c|}{15.40}
                                                                                 & 0.471                                   &0.972                    &0.469                       & \multicolumn{1}{c|}{15.36}
                                                                                 & 0.468                                   &0.969                    &0.517                       & 15.96     \\
\multicolumn{1}{c|}{}                          & \multicolumn{1}{c|}{SC (C)}     & \multicolumn{1}{c}{0.547}               &0.577                    &0.220                       & \multicolumn{1}{c|}{3.24}
                                                                                 & 0.177                                   &0.116                    &0.031                       & \multicolumn{1}{c|}{3.24}
                                                                                 & 0.373                                   &0.399                    &0.159                       & 3.29     \\
\multicolumn{1}{c|}{}                          & \multicolumn{1}{c|}{OT (G)}     & \multicolumn{1}{c}{0.740}               &0.964                    &0.708                       & \multicolumn{1}{c|}{\textbf{\bl{135.14}}}
                                                                                 & 0.651                                   &0.946                    &0.574                       & \multicolumn{1}{c|}{\textbf{\bl{133.51}}}
                                                                                 & 0.540                                   &0.958                    &0.553                       & \textbf{\bl{164.74}}     \\
\multicolumn{1}{c|}{}                          & \multicolumn{1}{c|}{RING++ (G)} & \multicolumn{1}{c}{\textbf{\bl{0.834}}} &\textbf{0.982}           &\textbf{\bl{0.831}}         & \multicolumn{1}{c|}{0.45}
                                                                                 & \textbf{0.701}                          &\textbf{\bl{0.993}}      &\textbf{0.729}              & \multicolumn{1}{c|}{0.44}
                                                                                 & \textbf{\bl{0.688}}                     &\textbf{0.974}           &\textbf{\bl{0.753}}         & 0.76    \\
\multicolumn{1}{c|}{}                          & \multicolumn{1}{c|}{Ours-BF (C)}& \multicolumn{1}{c}{\textbf{\gl{0.800}}}      &\textbf{\bl{0.987}}      &\textbf{0.830}              & \multicolumn{1}{c|}{\textbf{57.94}}
                                                                                 & \textbf{\bl{0.706}}                     &\textbf{\gl{0.990}}           &\textbf{\bl{0.746}}         & \multicolumn{1}{c|}{\textbf{57.34}}
                                                                                 & \textbf{\gl{0.639}}                          &\textbf{\bl{0.992}}      &\textbf{\gl{0.732}}              & \textbf{68.40}    \\
\multicolumn{1}{c|}{}                          & \multicolumn{1}{c|}{Ours-KD (C)}& \textbf{\gl{0.800}}                     &\textbf{\gl{0.986}}      &\textbf{\bl{0.831}}              & \multicolumn{1}{c|}{\gl{\textbf{\gl{82.58}}}}
                                                                                 & \textbf{\bl{0.706}}                          &\textbf{\gl{0.990}}      &\textbf{\gl{0.745}}              & \multicolumn{1}{c|}{\textbf{\gl{82.01}}}
                                                                                 & \textbf{\gl{0.639}}                          &\textbf{\bl{0.992}}      &\textbf{\gl{0.732}}              & \textbf{\gl{87.66}}     \\ \hline
\multicolumn{1}{c|}{\multirow{7}{*}{120}}      & \multicolumn{1}{c|}{Hist. (C)}  & \multicolumn{1}{c}{0.780}               &0.961                    &0.771                       & \multicolumn{1}{c|}{1.57}
                                                                                 & 0.483                                   &0.965                    &0.427                       & \multicolumn{1}{c|}{1.57}
                                                                                 & 0.596                                   &0.958                    &0.613                       & 1.61     \\
\multicolumn{1}{c|}{}                          & \multicolumn{1}{c|}{Logg3D (G)} & 0.746                                   &0.920                    &0.581                       & \multicolumn{1}{c|}{26.15} 
                                                                                 & 0.674                                   &0.983                    &0.664                       & \multicolumn{1}{c|}{26.03} 
                                                                                 & 0.625                                   &0.910                    &0.511                       & 28.14     \\
\multicolumn{1}{c|}{}                          & \multicolumn{1}{c|}{M2DP (C)}   & 0.751                                   &0.959                    &0.728                       & \multicolumn{1}{c|}{8.67}
                                                                                 & 0.404                                   &\textbf{0.987}           &0.441                       & \multicolumn{1}{c|}{8.66}
                                                                                 & 0.460                                   &\textbf{\gl{0.978}}           &0.525                       & 8.85     \\
\multicolumn{1}{c|}{}                          & \multicolumn{1}{c|}{SC (C)}     & 0.417                                   &0.393                    &0.152                       & \multicolumn{1}{c|}{3.21}
                                                                                 & 0.151                                   &0.155                    &0.045                       & \multicolumn{1}{c|}{3.20}
                                                                                 & 0.332                                   &0.379                    &0.126                       & 3.25     \\
\multicolumn{1}{c|}{}                          & \multicolumn{1}{c|}{OT (G)}     & 0.799                                   &0.927                    &0.705                       & \multicolumn{1}{c|}{\textbf{\bl{97.18}}}
                                                                                 & 0.691                                   &0.969                    &0.616                       & \multicolumn{1}{c|}{\textbf{\bl{96.33}}}
                                                                                 & 0.614                                   &0.891                    &0.516                       & \textbf{\bl{111.61}}     \\
\multicolumn{1}{c|}{}                          & \multicolumn{1}{c|}{RING++ (G)} & \textbf{\bl{0.862}}                     &\textbf{0.977}           &\textbf{\bl{0.880}}         & \multicolumn{1}{c|}{0.45}
                                                                                 & \textbf{\bl{0.750}}                     &0.978                    &\textbf{\bl{0.815}}         & \multicolumn{1}{c|}{0.44}
                                                                                 & \textbf{\bl{0.732}}                     &0.974                    &\textbf{\bl{0.793}}         & 0.75     \\
\multicolumn{1}{c|}{}                          & \multicolumn{1}{c|}{Ours-BF (C)}& \textbf{0.822}                          &\textbf{\bl{0.994}}      &\textbf{\gl{0.868}}              & \multicolumn{1}{c|}{\textbf{45.19}}
                                                                                 & \textbf{\gl{0.735}}                          &\textbf{\bl{0.992}}      &\textbf{0.790}              & \multicolumn{1}{c|}{\textbf{44.82}}
                                                                                 & \textbf{0.656}                          &\textbf{\bl{0.989}}      &\textbf{\gl{0.737}}              & \textbf{\gl{51.31}}     \\ 
\multicolumn{1}{c|}{}                          & \multicolumn{1}{c|}{Ours-KD (C)}& \textbf{\gl{0.840}}                                   &\textbf{\bl{0.994}}      &\textbf{0.867}              & \multicolumn{1}{c|}{\textbf{\gl{58.89}}}
                                                                                 & \textbf{\gl{0.735}}                          &\textbf{\bl{0.992}}      &\textbf{\gl{0.794}}              & \multicolumn{1}{c|}{\textbf{\gl{58.61}}}
                                                                                 & \textbf{\gl{0.659}}                          &\textbf{0.974}           &\textbf{0.710}              & \textbf{41.75}  \\ \hline 
\multicolumn{1}{c|}{\multirow{8}{*}{180}}       & \multicolumn{1}{c|}{Hist. (C)} & \multicolumn{1}{c}{0.780}               &0.966                    &0.761                       & \multicolumn{1}{c|}{1.09}
                                                                                 & 0.462                                   &0.965                    &0.410                       & \multicolumn{1}{c|}{1.08}
                                                                                 & 0.542                                   &0.952                    &0.570                       & 1.10     \\
\multicolumn{1}{c|}{}                          & \multicolumn{1}{c|}{Logg3D (G)} & 0.780                                   &0.918                    &0.609                       & \multicolumn{1}{c|}{22.87}
                                                                                 & 0.701                                   &0.966                    &0.647                       & \multicolumn{1}{c|}{22.77}
                                                                                 & 0.611                                   &0.903                    &0.492                       & 24.37     \\
\multicolumn{1}{c|}{}                          & \multicolumn{1}{c|}{M2DP (C)}   & 0.739                                   &0.940                    &0.694                       & \multicolumn{1}{c|}{6.03}
                                                                                 & 0.523                                   &0.981                    &0.555                       & \multicolumn{1}{c|}{6.02}
                                                                                 & 0.497                                   &0.921                    &0.526                       & 6.11     \\
\multicolumn{1}{c|}{}                          & \multicolumn{1}{c|}{SC (C)}     & 0.547                                   &0.577                    &0.694                       & \multicolumn{1}{c|}{3.16}
                                                                                 & 0.177                                   &0.116                    &0.031                       & \multicolumn{1}{c|}{3.16}
                                                                                 & 0.384                                   &0.353                    &0.146                       & 3.21     \\
\multicolumn{1}{c|}{}                          & \multicolumn{1}{c|}{OT (G)}     & \textbf{\gl{0.842}}                          &0.907                    &0.719                       & \multicolumn{1}{c|}{\textbf{\bl{76.34}}}
                                                                                 & 0.698                                   &0.962                    &0.646                       & \multicolumn{1}{c|}{\textbf{\bl{75.82}}}
                                                                                 & 0.645                                   &0.864                    &0.466                       & \textbf{\bl{84.96}}     \\
\multicolumn{1}{c|}{}                          & \multicolumn{1}{c|}{RING++ (G)} & \textbf{\bl{0.895}}                     &\textbf{0.976}           &\textbf{\bl{0.887}}         & \multicolumn{1}{c|}{0.45}
                                                                                 & \textbf{\bl{0.752}}                     &\textbf{0.983}           &\textbf{\bl{0.813}}         & \multicolumn{1}{c|}{0.44}
                                                                                 & \textbf{\bl{0.744}}                     &\textbf{\bl{0.988}}      &\textbf{\bl{0.795}}         & 0.74     \\ 
\multicolumn{1}{c|}{}                          & \multicolumn{1}{c|}{Ours-BF (C)}& \textbf{0.840}                                   &\textbf{\bl{0.994}}      &\textbf{0.868}              & \multicolumn{1}{c|}{\textbf{37.61}}
                                                                                 & \textbf{\gl{0.735}}                          &\textbf{\bl{0.990}}      &\textbf{0.799}              & \multicolumn{1}{c|}{\textbf{37.36}}
                                                                                 & \textbf{\gl{0.659}}                          &\textbf{\gl{0.974}}           &\textbf{\gl{0.710}}              & \textbf{41.75}  \\
\multicolumn{1}{c|}{}                          & \multicolumn{1}{c|}{Ours-KD (C)}& \textbf{0.840}                                    &\textbf{\bl{0.994}}      &\textbf{\gl{0.871}}              & \multicolumn{1}{c|}{\textbf{\gl{46.74}}}
                                                                                 & \textbf{\gl{0.735}}                          &\textbf{\bl{0.990}}      &\textbf{\gl{0.800}}              & \multicolumn{1}{c|}{\textbf{\gl{46.42}}}
                                                                                 & \textbf{\gl{0.659}}                          &\textbf{\gl{0.974}}           &\textbf{\gl{0.710}}              & \textbf{\gl{48.20}}  \\ \hline \bottomrule

\end{tabular}}
\vspace{-0.5cm}
\label{table:eval_pr}
\end{table*}
% ===================================================================================================================================================================================
\begin{table}[h]
\caption{Processing Time on Onboard Computer}
\renewcommand{\arraystretch}{1.5}
\centering\resizebox{0.48\textwidth}{!}{
\begin{tabular}{c|ccccccc}
\toprule \hline
\multirow{3}{*}{FOV} & \multicolumn{7}{c}{Processing Time (O) {[}s{]}}                                     \\ \cline{2-8} 
                     & \multicolumn{4}{c|}{CPU}        & \multicolumn{3}{c}{GPU}                                     \\ \cline{2-8} 
                     & Hist. (C)      & M2DP (C)                 & SC (C)          & \multicolumn{1}{c|}{Ours (C)}               & OT (G) & LoGG3D (G) & RING++ (G) \\ \hline
60                   & \textbf{1.289} & \gl{\textbf{0.297}}      & 1.359           & \multicolumn{1}{c|}{\bl{\textbf{0.073}}}    & 0.036  & 0.160      & 10.85    \\
120                  & 2.442          & \gl{\textbf{0.531}}      & \textbf{1.371}  & \multicolumn{1}{c|}{\bl{\textbf{0.093}}}    & 0.047  & 0.188      & 10.92    \\
180                  & 3.550          & \gl{\textbf{0.710}}      & \textbf{1.382}  & \multicolumn{1}{c|}{\bl{\textbf{0.111}}}    & 0.056  & 0.215      & 10.98    \\ \hline \bottomrule
                               % & 360                  & 6.996     & \textbf{0.312}      & 1.136    & 1.440  & \bl{\textbf{0.083}}  & 11.16      & \gl{\textbf{0.200}}    \\ \hline \bottomrule
\end{tabular}}
\vspace{-0.6cm}
\label{table:fov_time}
\end{table}
%% ===========================================================================================================================================================================
We also compared the proposed method using the kd-tree and brute force searching.
While the kd-tree is about 10-20Hz faster, its performance is slightly lower or similar.
Therefore, it can be effectively used according to the trade-off.
Unlike \tabref{table:eval_pr}, the \tabref{table:fov_time} reports processing time from the onboard computer in KITTI 00 sequence.
On the desktop, LoGG3D, M2DP, OT, and our method can operate in real-time, whereas on an onboard computer, only OT and the proposed method can achieve real-time.
However, OT falls short in terms of performance. 
Furthermore, as shown in \tabref{table:eval_pr} and \tabref{table:fov_time}, narrowing the FOV results in lower processing time consumption. 
This means that if we can robustly handle limited FOV, we can utilize descriptors more efficiently on the onboard computer.

\subsubsection{Rotational Robustness}
To evaluate the robustness of the rotation changes, we checked the top 1 recall by applying explicit heading changes on KITTI 05 with 180 FOV.
Except when the surroundings are completely symmetric, our method shows higher recall values compared to other methods.
%% ===========================================================================================================================================================================
\begin{figure}[h]
    \vspace{-0.2cm}
    \centering
    \def\width{1\columnwidth}
    \includegraphics[clip, trim= 120 70 120 110, width=\width]{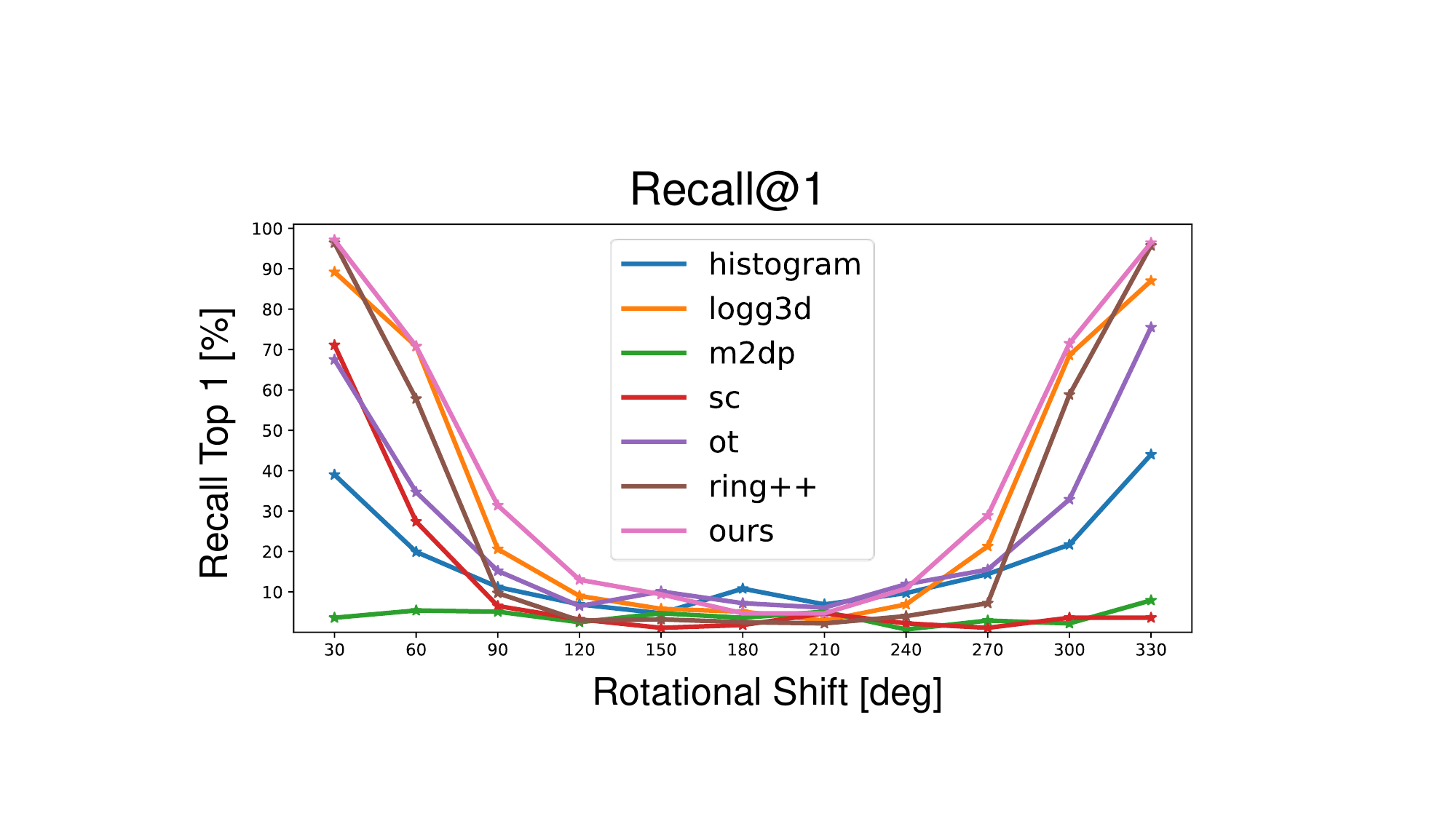}
    \vspace{-0.5cm}
    \caption{Experiment on KITTI 05 with a 180$^\circ$ FOV. The clipping angle of the query database is gradually shifted from front-facing to rotating, while the clipping angle of the candidate database remains constant front-facing.}
    \label{fig:rotational_robustness}
    \vspace{-0.2cm}
\end{figure} 
%% ===========================================================================================================================================================================

\noindent When the environment is completely symmetrical, it is difficult to recognize a revisited place with little information because the overlap is either non-existent or vertical planes.
On the other hand, when a rotational change occurs within a limited FOV, there is overlap and we can utilize this to counteract the rotational change as shown in \figref{fig:rotational_robustness}.

\subsubsection{Initial Heading Estimation} 
As represented in \figref{fig:main}, slight errors on narrow FOV can have a relatively larger impact compared to a 360$^\circ$ FOV.
Thus, we evaluated the RE between an estimated initial heading from \textit{A-SOLiD} and a ground truth heading.
\tabref{table:rot_error} indicates that \textit{A-SOLiD} estimates the initial heading to be around 1$^\circ$ of the ground truth heading, making it similar.
This means that our method can be given as the initial rotation for \ac{ICP}, which leads to accurate map building.
\begin{table}[h]
\vspace{-0.5cm}
\caption{Rotation Error}
\renewcommand{\arraystretch}{1.2}
\centering\resizebox{0.48\textwidth}{!}{
{\small
\begin{tabular}{c|ccc|ccc|ccc}
\toprule
\hline
Sequence     & \multicolumn{3}{c|}{KITTI 00} & \multicolumn{3}{c|}{KITTI 02} & \multicolumn{3}{c}{KITTI 05} \\ \hline
FOV          & \multicolumn{1}{c|}{60} & \multicolumn{1}{c|}{90} & 180 & \multicolumn{1}{c|}{60} & \multicolumn{1}{c|}{90} & 180 & \multicolumn{1}{c|}{60} & \multicolumn{1}{c|}{90} & 180 \\ \hline
RE {[}$^\circ${]} & \multicolumn{1}{c|}{0.24}   & \multicolumn{1}{c|}{0.76}   & 0.77    & \multicolumn{1}{c|}{0.59}   & \multicolumn{1}{c|}{0.69}   & 0.99    & \multicolumn{1}{c|}{0.72}   & \multicolumn{1}{c|}{0.88}   & 1.07    \\ \hline \bottomrule
\end{tabular}}}
\vspace{-0.4cm}
\label{table:rot_error}
\end{table}

\subsubsection{Occulsion} 
In robot navigation, occlusion can often be encountered when there are nearby objects.
Assuming such a situation, we restricted the FOV as shown in \figref{fig:various_fov}.
%% ROC curve
%% ===========================================================================================================================================================================
\vspace{-0.4cm}
\begin{figure}[h]
    \centering
    \def\width{1\columnwidth}
    \includegraphics[clip, trim= 60 635 60 10, width=\width]{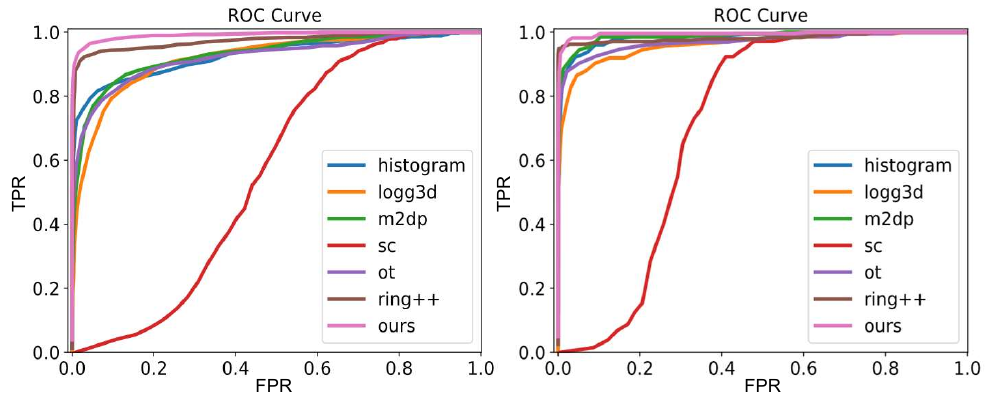}
    \vspace{-0.8cm}
    \caption{The ROC curves when LiDAR scan is occluded. Left is KITTI 00's evaluation, and right is KITTI 02's evaluation.}
    \label{fig:roc_curve}
    \vspace{-0.4cm}
\end{figure} 
%% ===========================================================================================================================================================================

\noindent As represented in \figref{fig:roc_curve}, our method's graph is positioned at the highest point.
In other words, our method can reduce drift error in occlusion situations through effective loop distinction.

%% Matching Graph
%% ===========================================================================================================================================================================
\begin{figure*}[t]
    \centering
    \def\width{2\columnwidth}
    \includegraphics[clip, trim= 0 268 0 65, width=\width]{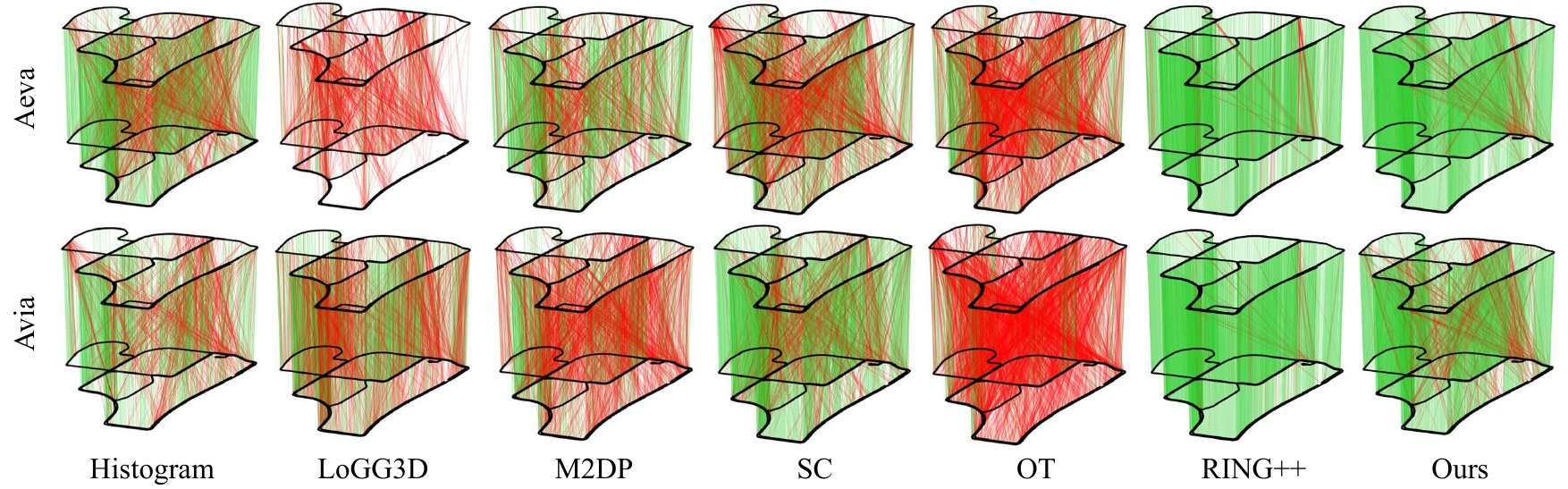}
    \vspace{-0.3cm}
    \caption{3-D matching pair graph when each descriptor's f1 score is highest in HeLiPR dataset (source: KAIST04, target: KAIST05). Green represents true matching pairs, and red represents false matching pairs.}
    \vspace{-0.6cm}
    \label{fig:matching_Graph}
\end{figure*}
%% ===========================================================================================================================================================================
\vspace{-0.4cm}
\subsection{Multi session Place Recognition}
\begin{figure}[t]
    \centering
    \def\width{1\columnwidth}
    \includegraphics[clip, trim= 0 340 0 20, width=\width]{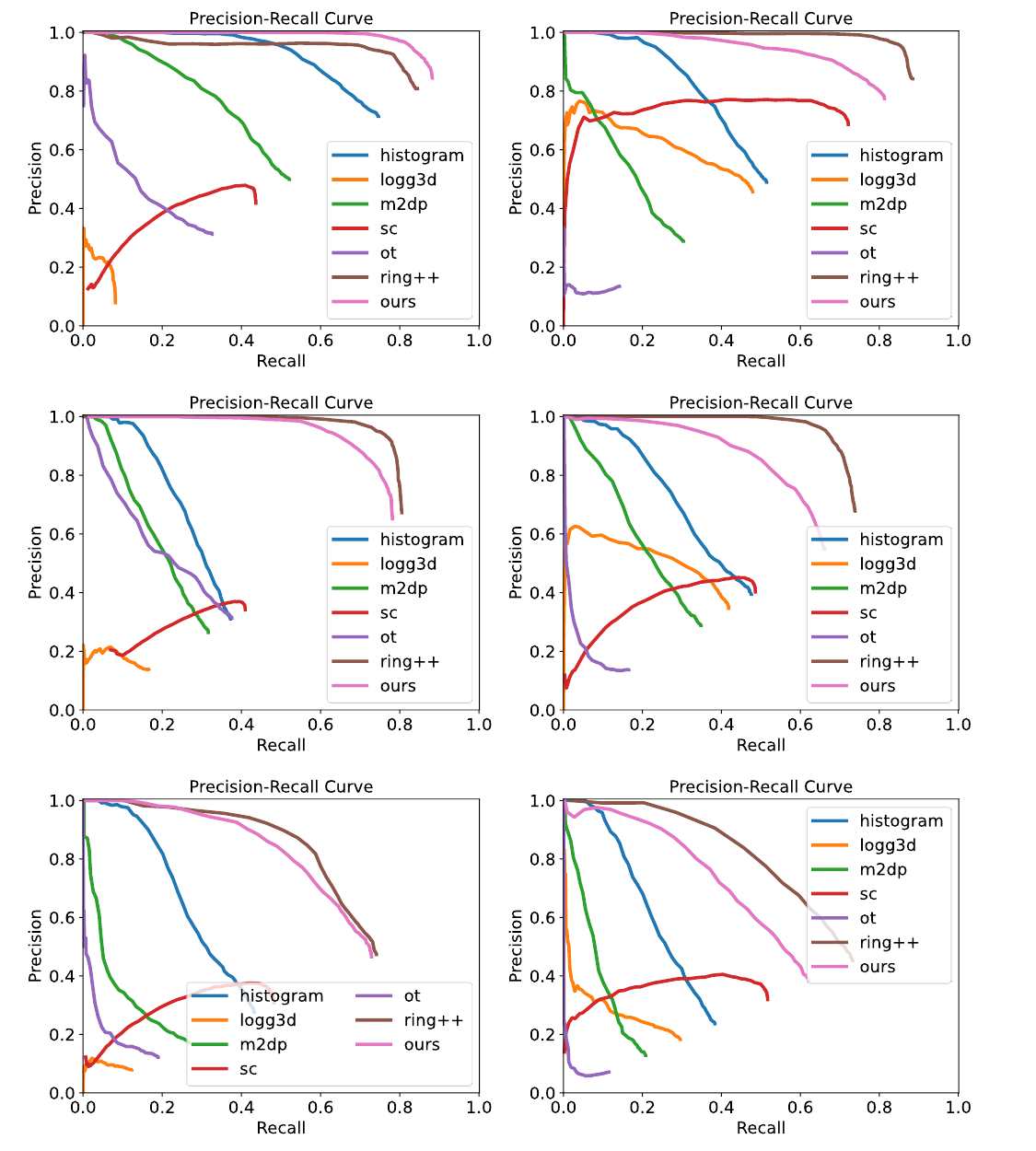}
    \vspace{-0.8cm}
    \caption{PR curve on HeLiPR datasets. The left column used Aeva LiDAR and the right used Avia LiDAR.
             The top represents KAIST04-05, the middle means Town01-02 and the bottom means RoundAbout01-02 session.}
    \vspace{-0.5cm}
    \label{fig:pr_curve}
\end{figure}

We evaluated multi-session scenarios for long-term autonomy.
As shown in \figref{fig:pr_curve}, the proposed method, Histogram, and RING++ can maintain a high precision at a consistent recall level.
Thus, these three methods possess the reliable ability to distinguish between false loops and true loops.
\figref{fig:matching_Graph} shows a 3-D loop matching graph in KAIST04-05 of multi-session, acquired with solid-state LiDAR.
With Aeva, our method finds the many loops, while in Avia, RING++ shows the highest performance.
\tabref{table:matching_number} refers to the number of true positive and false positive loops in the \figref{fig:matching_Graph}'s 3-D graph.
%% PR curve
%% ===========================================================================================================================================================================
%% ===========================================================================================================================================================================
In the Aeva sequence, RING++ achieves a precision that is about 0.03 higher. 
However, our method discovers higher recall and consequently maintains a higher f1 score.
Especially when precision is 100$\%$, our method detects more loops as shown in Appendix.
In the Avia sequence, RING++ demonstrates superiority. 
However, it necessitates the use of GPU as indicated in the \tabref{table:helipr_time}, leading to longer processing times.
%% Matching Number
%% ===========================================================================================================================================================================
\begin{table}[h]
\vspace{-0.3cm}
\caption{Number of Matching Pair in KAIST04-05}
\renewcommand{\arraystretch}{1.5}
\centering\resizebox{0.48\textwidth}{!}{
\begin{tabular}{c|c|ccccccc}
\toprule \hline
LiDAR Type            & Metric            & Hist. (C)      & LoGG3D (G)  & M2DP (C)     & SC (C)         & OT (G)  & RING++ (G)           & Ours (C) \\ \hline
\multirow{3}{*}{Aeva} & TP $\uparrow$     & \textbf{1526}  & 200         & 1191         & 1155           & 874     & \gl{\textbf{2116}}   & \bl{\textbf{2291}}  \\
                      & FP $\downarrow$   & 848            & 884         & \textbf{740} & 1314           & 1888    & \bl{\textbf{171}}    & \gl{\textbf{196}}     \\ 
                      & F1 max $\uparrow$ & \textbf{0.603} & 0.106       & 0.516        & 0.448          & 0.321   & \gl{\textbf{0.851}}  & \bl{\textbf{0.886}} \\ \hline
\multirow{3}{*}{Avia} & TP $\uparrow$     & 884            & 1233        & 732          & \textbf{1887}  & 379     & \bl{\textbf{2208}}   & \gl{\textbf{2119}}         \\
                      & FP $\downarrow$   & \textbf{607}   & 1271        & 1508         & 717            & 2424    & \bl{\textbf{109}}    & \gl{\textbf{501}}     \\ 
                      & F1 max $\uparrow$ & 0.424          & 0.477       & 0.298        & \textbf{0.716} & 0.138   & \bl{\textbf{0.901}}  & \gl{\textbf{0.801}}    \\ 
\hline
\bottomrule
\end{tabular}}
\label{table:matching_number}
% \vspace{-0.35cm}
\end{table}
%% ===========================================================================================================================================================================
%% Processing Time
%% ===========================================================================================================================================================================
\begin{table}[h]
\vspace{-0.15cm}
\caption{Processing Time in solid-state LiDAR}
\renewcommand{\arraystretch}{1.5}
\centering\resizebox{0.48\textwidth}{!}{
\begin{tabular}{c|c|ccccccc}
\toprule \hline
\multirow{3}{*}{Computer}    & \multirow{3}{*}{LiDAR Type} & \multicolumn{7}{c}{Processing Time {[}s{]}}                                          \\ \cline{3-9} 
                             &                             & \multicolumn{4}{c|}{CPU}        & \multicolumn{3}{c}{GPU}                            \\  \cline{3-9}
                             &     & Hist. (C) & M2DP (C)  & SC (C)  & \multicolumn{1}{c|}{Ours (C)}                  & OT (G)              & LoGG3D (G) & RING++ (G)  \\  \hline
\multirow{2}{*}{O}
                                 
                            & Aeva & 5.291          & \gl{\textbf{1.109}}     & \textbf{1.439}   & \multicolumn{1}{c|}{\bl{\textbf{0.143}}}       & 0.083               & 0.229          & 15.70    \\
                            & Avia & \textbf{1.260} & \gl{\textbf{0.258}}     & 1.433            & \multicolumn{1}{c|}{\bl{\textbf{0.137}}}       & 0.047               & 0.189          & 15.58   \\ \hline
\multirow{2}{*}{D}         
                                 
                            & Aeva & 1.426     & 0.271                   & 0.145            & \multicolumn{1}{c|}{\gl{\textbf{0.036}}}       & \bl{\textbf{0.021}} & \textbf{0.056} & 3.720    \\
                            & Avia & 0.323     & 0.072                   & 0.142            & \multicolumn{1}{c|}{\gl{\textbf{0.033}}}       & \bl{\textbf{0.012}} & \textbf{0.039} & 3.680    \\ \hline\bottomrule
\end{tabular}}
\label{table:helipr_time}
\vspace{-0.3cm}
\end{table}
%% ===========================================================================================================================================================================
Also, OT has a short processing time but lacks in terms of f1 score or performance.
On the other hand, our method offers both a short processing time and superior performance. 
Also, by utilizing kd-tree, even faster speeds can be achieved.

\vspace{-0.4cm}
\subsection{Multi-robot Place Recognition}
In multi-robot systems, the communication time to exchange descriptors for quickly and accurately knowing the positions of each other's robots is crucial.
However, there are various constraints on communication, such as limited bandwidth in real multi-robot mapping. 
Therefore, within the given time, we need to transmit and receive important information. 
Longer communication times can lead to bottlenecks and increase the likelihood of data loss.

To measure communication time, we followed three steps.
First, we assumed that each robot was installed on an onboard computer as represented in \tabref{table:setup}.
Second, we measured each onboard computer's bandwidth under the assumption that there is no limitation on communication distance.
Finally, we derived the communication time as follows:
\begin{equation}
\text { Communication Time } =\text { Description } \times \frac{1}{\text {Bandwidth}},
\end{equation}
where the communication time, description, and bandwidth have units of [s], [B], and [B/s], respectively.
\begin{table}[h]
\vspace{-0.25cm}
\caption{Hardware setup and Data size}
\renewcommand{\arraystretch}{1.2}
\centering\resizebox{0.48\textwidth}{!}{
{\huge
\begin{tabular}{c|c|ccccccc}
\toprule
\hline
\multirow{2}{*}{Robot ID} & \multirow{2}{*}{Hardware} & \multicolumn{7}{c}{Data size [MB] }                                               \\ \cline{3-9} 
                          &                           & Hist. (C)          & LoGG3D (G) & M2DP (C) & SC (C) & OT (G)          & RING++ (G) & Ours (C) \\ \hline
Robot 1                   & Jetson AGX Orin           & \gl{\textbf{0.5}}  & 1.3        & 1.0      & 6.0    & \textbf{0.7}    & 202        & \bl{\textbf{0.3}}      \\
Robot 2                   & Intel NUC 10 Mini PC      & \gl{\textbf{0.6}}  & 1.5        & 1.4      & 7.0    & \textbf{0.8}    & 242        & \bl{\textbf{0.3}}      \\
Robot 3                   & Intel NUC 12 Pro Mini PC  & \gl{\textbf{0.8}}  & 1.8        & 1.4      & 8.0    & \textbf{1.0}    & 294        & \bl{\textbf{0.4}}      \\ \hline
\bottomrule
\end{tabular}}}
\label{table:setup}
\vspace{-0.3cm}
\end{table}

We assumed that all descriptors generated among the three robots are exchanged after the mapping is completed.
Thus, we evaluated the final result based on the sum of the total six communication times.
As shown in \tabref{table:dist_eval}, our method exhibits the shortest communication speed.
On the other hand, RING++ exhibits a relatively longer communication time. 
% RING++ employs a 3-D descriptor of size 6$\times$120$\times$120, as it minimizes compression in the 3-D space.
% While this lack of dimension reduction allows for a more diverse representation of the scene, it is natively a heavyweight descriptor, as indicated in the \tabref{table:description}.
% From this perspective, RING++ is approximately 100$\times$ more likely to experience exchange loss compared to the proposed method.
% Therefore, our method allows for robust and efficient multi-robot operation without significant concern for network bottlenecks.
Also, we evaluated the overall PR performance in multi-robot sequences with narrow FOV. 
As described in \tabref{table:dist_eval}, the performance is superior in the order of RING++, the proposed method, and LoGG3D or M2DP. 
Although our method has slightly inferior performance compared to RING++, short communication time makes the proposed method more efficient for multi-robot scenarios.

%% ===========================================================================================================================================================================
\begin{table}[h]
\vspace{-0.15cm}
\caption{Evaluation for Multi-Robot System}
\centering\resizebox{0.48\textwidth}{!}{
\begin{tabular}{c|c|ccc|c}
\toprule \hline
Method                     & FOV                      & Recall@1$\uparrow$   & AUC$\uparrow$       & F1 max$\uparrow$     & Commu. Time $\downarrow$ \\ \hline
\multirow{3}{*}{Hist. (C)} & \multicolumn{1}{c|}{60}  & 0.405                & 0.847               & 0.349                & \multirow{3}{*}{\textbf{\gl{0.285s}}}          \\
                           & \multicolumn{1}{c|}{120} & 0.431                & 0.862               & 0.373                &                            \\
                           & \multicolumn{1}{c|}{180} & 0.465                & 0.856               & 0.375                &                            \\ \hline
\multirow{3}{*}{LoGG3D (G)}& \multicolumn{1}{c|}{60}  & \textbf{\gl{0.590}}  & 0.843               & \textbf{0.387}       & \multirow{3}{*}{0.669s}          \\
                           & \multicolumn{1}{c|}{120} & 0.647                & 0.890               & \textbf{0.471}       &                            \\
                           & \multicolumn{1}{c|}{180} & 0.654                & \textbf{0.910}      & \textbf{0.516}       &                            \\ \hline
\multirow{3}{*}{M2DP (C)}  & \multicolumn{1}{c|}{60}  & 0.379                & \textbf{0.874}      & 0.359                & \multirow{3}{*}{0.511s}          \\
                           & \multicolumn{1}{c|}{120} & 0.433                & \textbf{\gl{0.919}} & 0.417                &                            \\
                           & \multicolumn{1}{c|}{180} & 0.432                & 0.888               & 0.383                &                            \\ \hline
\multirow{3}{*}{SC (C)}    & \multicolumn{1}{c|}{60}  & 0.474                & 0.648               & 0.272                & \multirow{3}{*}{3.077s}          \\
                           & \multicolumn{1}{c|}{120} & 0.468                & 0.579               & 0.272                &                            \\
                           & \multicolumn{1}{c|}{180} & 0.497                & 0.570               & 0.278                &                            \\ \hline
\multirow{3}{*}{OT (G)}    & \multicolumn{1}{c|}{60}  & 0.366                & 0.867               & 0.348                & \multirow{3}{*}{\textbf{0.354s}}          \\
                           & \multicolumn{1}{c|}{120} & 0.423                & 0.834               & 0.369                &                            \\
                           & \multicolumn{1}{c|}{180} & 0.482                & 0.864               & 0.395                &                            \\ \hline
\multirow{3}{*}{RING++ (G)}& \multicolumn{1}{c|}{60}  & \textbf{\bl{0.670}}  & \textbf{\bl{0.940}} & \textbf{\bl{0.585}}  & \multirow{3}{*}{106.221s}          \\
                           & \multicolumn{1}{c|}{120} & \textbf{\bl{0.701}}  & \textbf{\bl{0.975}} & \textbf{\bl{0.730}}  &                            \\
                           & \multicolumn{1}{c|}{180} & \textbf{\bl{0.707}}  & \textbf{\bl{0.986}} & \textbf{\bl{0.767}}  &                             \\ \hline
\multirow{3}{*}{Ours (C)}  & \multicolumn{1}{c|}{60}  & \textbf{0.572}       & \textbf{\gl{0.881}} & \textbf{\gl{0.442}}  & \multirow{3}{*}{\textbf{\bl{0.138s}}}          \\
                           & \multicolumn{1}{c|}{120} & \textbf{\gl{0.665}}  & \textbf{\gl{0.919}} & \textbf{\gl{0.548}}  &                            \\
                           & \multicolumn{1}{c|}{180} & \textbf{\gl{0.689}}  & \textbf{\gl{0.930}} & \textbf{\gl{0.571}}  &                            \\ \hline \bottomrule
\end{tabular}}
\label{table:dist_eval}
\vspace{-0.5cm}
\end{table}
%% ===========================================================================================================================================================================

\subsection{Ablation Study}
The key components of the loop detection are $N_r$ and~$N_e$.
These values determined the \textit{REC} of spatial organization, and we conducted an ablation study to investigate its impact.
\begin{table}[h]
\vspace{-0.15cm}
\caption{Ablation Study for Loop Detection}
\renewcommand{\arraystretch}{1.2}
\centering\resizebox{0.45\textwidth}{!}{
\begin{tabular}{c|c|cccccc}
\toprule \hline
\multirow{2}{*}{Variable} & \multirow{2}{*}{Evaluation} & \multicolumn{6}{c}{Value}                     \\ \cline{3-8} 
                          &             & 20                  & 40                  & 60                    & 80                  & 100                 & 120   \\ \hline
\multirow{3}{*}{$N_r$}    & Recall@1    & \textbf{0.871}      & \textbf{\bl{0.874}} & 0.869                 & 0.869               & \textbf{\gl{0.873}} & \textbf{0.871} \\ \cline{2-8}
                          & AUC         & \textbf{\gl{0.996}} & \textbf{\gl{0.996}} & \textbf{\bl{0.997}}   & \textbf{\gl{0.996}}               & 0.995               & 0.994 \\ \cline{2-8}
                          & F1 max      & \textbf{0.887}      & \textbf{\bl{0.894}} & \textbf{\gl{0.891}}   & 0.883               & 0.877               & 0.866 \\ \hline
\multirow{3}{*}{$N_e$}    & Recall@1    & \textbf{\gl{0.872}} & \textbf{\bl{0.873}} & 0.871                 & \textbf{\gl{0.872}} & 0.866               & 0.871 \\ \cline{2-8}
                          & AUC         & \textbf{\gl{0.996}} & \textbf{\gl{0.996}} & \textbf{\gl{0.996}}   & \textbf{\gl{0.996}} & \textbf{\bl{0.997}} & \textbf{\gl{0.996}} \\  \cline{2-8}
                          & F1 max      & \textbf{0.892}      & \textbf{0.892}      & \textbf{\gl{0.893}}   & \textbf{0.892}      & \textbf{0.892}      & \textbf{\bl{0.894}} \\ \hline \bottomrule
\end{tabular}}
\label{table:abs_rsolid}
\vspace{-0.3cm}
\end{table}

\noindent \tabref{table:abs_rsolid} illustrates the LiDAR \ac{PR} performance of \textit{R-SOLiD} as it varies with $N_r$ and $N_e$ in KITTI 00.
While there is some variation based on the values, it can be observed that the changes are negligible, indicating that heuristic efforts through parameter tuning are not crucial.

%% ===========================================================================================================================================================================

%% file: PaperWriting/conclusion.tex
\section{Conclusion}
We proposed a spatially organized and lightweight global descriptor for FOV-constrained LiDAR \ac{PR}.
To address limited FOV scenarios, we effectively represented the scene through spatial organization and reweighting.
Furthermore, we confirmed efficient operation within onboard computing via lightweight description and fast searching. 
However, most LiDAR place recognition methods, including ours, show weakness in detecting places with reverse directions of limited FOV scenarios. 
% When a robot navigates in the opposite direction to a revisited place, an overlapping region of the LiDAR views between the current place and the revisited place may not occur, or only a small part of the vertical plane may overlap, even if the translation difference is a little apart.
If the robot is traveling in the opposite direction from the revisited place, there may be no overlap in the LiDAR view between the current place and the revisited place, or only a part of the vertical plane may overlap, even if the difference in translation is small.
Therefore, to recognize the reverse revisited place with all methods, including our descriptor, it is necessary to leverage 360$^\circ$ information or utilize additional information, such as semantics or sequential queues.
The applicability of our descriptor in the 360$^\circ$ view is confirmed as described in the Appendix on the supplementary materials' page.

In future works, we plan to build up sequential information to create a descriptor for recognizing revisited places in the opposite direction with a limited FOV. 
Also, we will integrate our method with the LiDAR odometry approaches and release a distributed SLAM framework.